\documentclass[10pt]{article}
\usepackage[letterpaper,top=0.78in,bottom=0.82in,left=0.88in,right=0.88in]{geometry}
\usepackage[T1]{fontenc}
\usepackage[utf8]{inputenc}
\usepackage{newtxtext}
\usepackage{amsmath,amsthm,mathtools}
\usepackage{newtxmath}
\usepackage[scaled=0.92]{helvet}
\usepackage[varqu]{inconsolata}
\usepackage{aliascnt}
\usepackage{bm}
\usepackage{booktabs,longtable,array,multirow,tabularx}
\usepackage{graphicx}
\usepackage{tikz}
\usetikzlibrary{arrows.meta,positioning,fit,calc,backgrounds,shapes.geometric,decorations.pathreplacing,matrix}
\usepackage{xcolor}
\usepackage{microtype}
\usepackage{enumitem}
\usepackage{caption}
\usepackage{float}
\usepackage[numbers,sort&compress]{natbib}
\usepackage{titlesec}
\usepackage{xurl}
\usepackage{placeins}
\usepackage{setspace}
\usepackage[hidelinks]{hyperref}
\usepackage{doi}
\usepackage[nameinlink,capitalise,noabbrev]{cleveref}

\definecolor{FFBlue}{HTML}{4C78A8}
\definecolor{FFBlueLight}{HTML}{EAF1F8}
\definecolor{FFGold}{HTML}{D3A52F}
\definecolor{FFGoldLight}{HTML}{FBF4DF}
\definecolor{FFRed}{HTML}{B56470}
\definecolor{FFRedLight}{HTML}{F8ECEE}
\definecolor{FFGray}{HTML}{6A7179}
\definecolor{FFGrayDark}{HTML}{3F454B}
\definecolor{FFRule}{HTML}{D3D7DB}
\definecolor{FFLight}{HTML}{F7F8F9}
\definecolor{FFInk}{HTML}{111315}

\color{FFInk}
\setlist{leftmargin=*,topsep=3pt,itemsep=1.8pt,parsep=0pt,partopsep=0pt}
\newcolumntype{L}[1]{>{\raggedright\arraybackslash}p{#1}}
\newcolumntype{Y}{>{\raggedright\arraybackslash}X}
\hypersetup{
  pdfauthor={Qinyou Wang},
  pdftitle={Fiber Fingerprints of Hidden Learning-State Dynamics},
  pdfsubject={Predictive quotients, history-reachable completion, chronology, and optimizer revelation},
  pdfkeywords={learning state, predictive quotient, fiber fingerprint, hidden learning dynamics, Transformer, AdamW, observability},
  pdfcreator={LaTeX with TikZ/PGF and Matplotlib},
  pdfdisplaydoctitle=true
}
\titleformat{\section}{\large\bfseries}{\thesection}{0.68em}{}
\titleformat{\subsection}{\normalsize\bfseries}{\thesubsection}{0.60em}{}
\titleformat{\subsubsection}{\normalsize\bfseries\itshape}{\thesubsubsection}{0.52em}{}
\titlespacing*{\section}{0pt}{12pt plus 3pt minus 2pt}{4.5pt plus 1pt minus 1pt}
\titlespacing*{\subsection}{0pt}{8.5pt plus 2pt minus 1pt}{2.8pt plus 1pt minus 1pt}
\titlespacing*{\subsubsection}{0pt}{6.5pt plus 1.5pt minus 1pt}{2.2pt}

\newtheoremstyle{ffplain}{6pt}{6pt}{\itshape}{}{\bfseries}{.}{0.5em}{}
\newtheoremstyle{ffdefinition}{6pt}{6pt}{\normalfont}{}{\bfseries}{.}{0.5em}{}
\newtheoremstyle{ffremark}{5.5pt}{5.5pt}{\normalfont}{}{\itshape}{.}{0.5em}{}
\theoremstyle{ffplain}
\newtheorem{theorem}{Theorem}[section]
\newaliascnt{proposition}{theorem}
\newtheorem{proposition}[proposition]{Proposition}
\aliascntresetthe{proposition}
\newaliascnt{corollary}{theorem}
\newtheorem{corollary}[corollary]{Corollary}
\aliascntresetthe{corollary}
\newaliascnt{lemma}{theorem}

\aliascntresetthe{lemma}
\theoremstyle{ffdefinition}
\newaliascnt{definition}{theorem}
\newtheorem{definition}[definition]{Definition}
\aliascntresetthe{definition}
\newaliascnt{assumption}{theorem}
\newtheorem{assumption}[assumption]{Assumption}
\aliascntresetthe{assumption}
\newaliascnt{counterexample}{theorem}
\newtheorem{counterexample}[counterexample]{Counterexample}
\aliascntresetthe{counterexample}
\newaliascnt{boundary}{theorem}
\newtheorem{boundary}[boundary]{Boundary}
\aliascntresetthe{boundary}
\theoremstyle{ffremark}
\newaliascnt{remark}{theorem}
\newtheorem{remark}[remark]{Remark}
\aliascntresetthe{remark}

\makeatletter
\renewcommand{\maketitle}{%
  \begin{center}
    {\fontsize{19}{22.5}\selectfont\bfseries \@title\par}
    \vspace{1.15em}
    {\normalsize \@author\par}
  \end{center}
  \vspace{0.50em}
}
\makeatother

\renewenvironment{abstract}{%
  \begin{center}\begin{minipage}{0.92\linewidth}\small
  \begin{center}\bfseries Abstract\end{center}\vspace{-0.45em}
}{%
  \end{minipage}\end{center}\vspace{0.35em}
}

\tikzset{
  ffbox/.style={
    draw=FFGrayDark,
    line width=0.66pt,
    rounded corners=1.1pt,
    align=center,
    inner xsep=2.7mm,
    inner ysep=1.8mm,
    font=\sffamily\footnotesize,
    text=FFInk
  },
  ffneutral/.style={ffbox,fill=FFLight},
  ffblue/.style={ffbox,fill=FFBlueLight},
  ffgold/.style={ffbox,fill=FFGoldLight},
  ffred/.style={ffbox,fill=FFRedLight},
  ffwhite/.style={ffbox,fill=white},
  ffgray/.style={ffbox,fill=FFLight,draw=FFRule},
  ffdashframe/.style={
    draw=FFGray,
    line width=0.58pt,
    rounded corners=1.2pt,
    dashed,
    fill=white,
    inner sep=2.4mm,
    align=center,
    font=\sffamily\footnotesize,
    text=FFInk
  },
  ffarrow/.style={
    -{Latex[length=1.85mm,width=1.12mm]},
    draw=FFGrayDark,
    line width=0.70pt,
    rounded corners=1.0pt
  },
  ffarrowblue/.style={
    -{Latex[length=1.85mm,width=1.12mm]},
    draw=FFBlue,
    line width=0.78pt,
    rounded corners=1.0pt
  },
  ffarrowgold/.style={
    -{Latex[length=1.85mm,width=1.12mm]},
    draw=FFGold,
    line width=0.78pt,
    rounded corners=1.0pt
  },
  ffarrowred/.style={
    -{Latex[length=1.85mm,width=1.12mm]},
    draw=FFRed,
    line width=0.78pt,
    rounded corners=1.0pt
  },
  ffdasharrow/.style={
    -{Latex[length=1.72mm,width=1.04mm]},
    draw=FFGray,
    line width=0.58pt,
    dashed,
    rounded corners=1.0pt
  },
  ffline/.style={draw=FFGrayDark,line width=0.64pt},
  ffrule/.style={draw=FFRule,line width=0.58pt},
  ffdashline/.style={draw=FFGray,line width=0.56pt,dashed},
  ffpanel/.style={font=\sffamily\scriptsize\bfseries,text=FFGrayDark,anchor=west},
  fflabel/.style={font=\sffamily\scriptsize,text=FFInk,align=center},
  ffsubtle/.style={font=\sffamily\tiny,text=FFGray,align=center},
  ffmath/.style={font=\footnotesize,text=FFInk,align=center},
  ffdot/.style={circle,minimum size=2.5pt,inner sep=0pt,draw=none,fill=FFGrayDark}
}

\crefname{theorem}{Theorem}{Theorems}
\Crefname{theorem}{Theorem}{Theorems}
\crefname{proposition}{Proposition}{Propositions}
\Crefname{proposition}{Proposition}{Propositions}
\crefname{corollary}{Corollary}{Corollaries}
\Crefname{corollary}{Corollary}{Corollaries}
\crefname{lemma}{Lemma}{Lemmas}
\Crefname{lemma}{Lemma}{Lemmas}
\crefname{definition}{Definition}{Definitions}
\Crefname{definition}{Definition}{Definitions}
\crefname{assumption}{Assumption}{Assumptions}
\Crefname{assumption}{Assumption}{Assumptions}
\crefname{counterexample}{Counterexample}{Counterexamples}
\Crefname{counterexample}{Counterexample}{Counterexamples}
\crefname{remark}{Remark}{Remarks}
\Crefname{remark}{Remark}{Remarks}
\crefname{boundary}{Boundary}{Boundaries}
\Crefname{boundary}{Boundary}{Boundaries}

\numberwithin{equation}{section}
\newcommand{\FF}{\mathsf{FF}}
\newcommand{\Birr}{\mathfrak B^{\mathrm{irr}}}

\newcommand{\E}{\mathbb E}

\newcommand{\cH}{\mathcal H}

\newcommand{\cP}{\mathcal P}
\newcommand{\cC}{\mathcal C}

\newcommand{\cU}{\mathcal U}

\newcommand{\cY}{\mathcal Y}
\newcommand{\Id}{\mathrm{Id}}
\newcommand{\Ran}{\operatorname{Ran}}
\newcommand{\Ker}{\operatorname{Ker}}
\newcommand{\rank}{\operatorname{rank}}
\newcommand{\tr}{\operatorname{tr}}
\newcommand{\diag}{\operatorname{diag}}

\newcommand{\HS}{\mathrm{HS}}
\newcommand{\op}{\mathrm{op}}

\title{Fiber Fingerprints of Hidden Learning-State Dynamics}
\author{Qinyou Wang}
\date{}

\begin{document}
\maketitle
\thispagestyle{empty}
\begin{abstract}
A learning system can occupy execution states that are indistinguishable under every declared present-behavior readout yet respond differently to future training. We formalize this through \emph{fiber fingerprints}: controlled future-learning response laws restricted to present-behavior equivalence classes. Prefix-compatible finite probes induce a predictive quotient functor, a Nerode-type minimal recursively sufficient representation, and a canonical set-level predictive fiber without assuming smoothness, reversibility, finite rank, or a manifold. Under an explicit finite-dimensional Hilbert realization, response decomposes into visible, visible-mode-reuse, and irreducible-new sectors; a history-reachability bridge retains only distinctions generated by natural training histories. Conditional mechanism results then identify a graph-Hodge chronology decomposition, a regular switching class with root-mean-square scale $\sqrt p\,\eta^{3/2}$ and finite-scale corrections, and an exact Adam moment section whose immediate adaptive field is constant while common future gradients can reveal hidden moment differences. Frozen Transformer--LoRA--AdamW studies with Qwen2.5-7B and Mistral-7B-v0.3 support a local action backbone, longer-horizon first-return non-closure, and fresh visible-relative completion with output-range reuse and a low-rank irreducible sector. Stronger claims remain bounded by preregistered negative or mixed results: re-anchored transport is unresolved above its measurement floor; the strict finite-grid Hodge--$3/2$ conjunction is unmet despite prospective contraction; Qwen accessibility is not established in the frozen raw moment chart; and Mistral revelation is future-context dependent rather than bank invariant. Within these support-, scale-, metric-, and context-resolved boundaries, present behavior is not a sufficient statistic for declared future learning.
\end{abstract}

\section{Introduction}
\label{sec:intro}

A model checkpoint is commonly summarized by present loss, benchmark accuracy, or an output distribution. A learning system, however, is an \emph{execution state}: parameters, optimizer moments, step counters, schedules, random streams, data position, accumulation buffers, precision state, and backend variables can all alter the next training trajectory. The question studied here is therefore not whether two parameter vectors compute the same present function, but whether two execution states that are indistinguishable under a declared present readout must learn in the same way.

They need not. Function-preserving ReLU rescalings can substantially change optimization dynamics \citep{lebeurrier2026path}; hidden positive-homogeneous gauges can determine feature specialization even when the complete initial predictor is fixed \citep{wang2026hiddengauge}; and low-rank recurrent networks can contain loss-invisible overlaps that encode training history and are revealed by learning \citep{ger2026invisible}. Optimizer memory and local task-order geometry likewise create structured order effects in fine-tuning \citep{sweeney2026memory,piontkovskaia2026fragile}. These results establish that present function or loss can omit learning-relevant state. The contribution here is an operational state theory that makes present equivalence explicit and uses controlled future training itself as the distinguishing experiment.

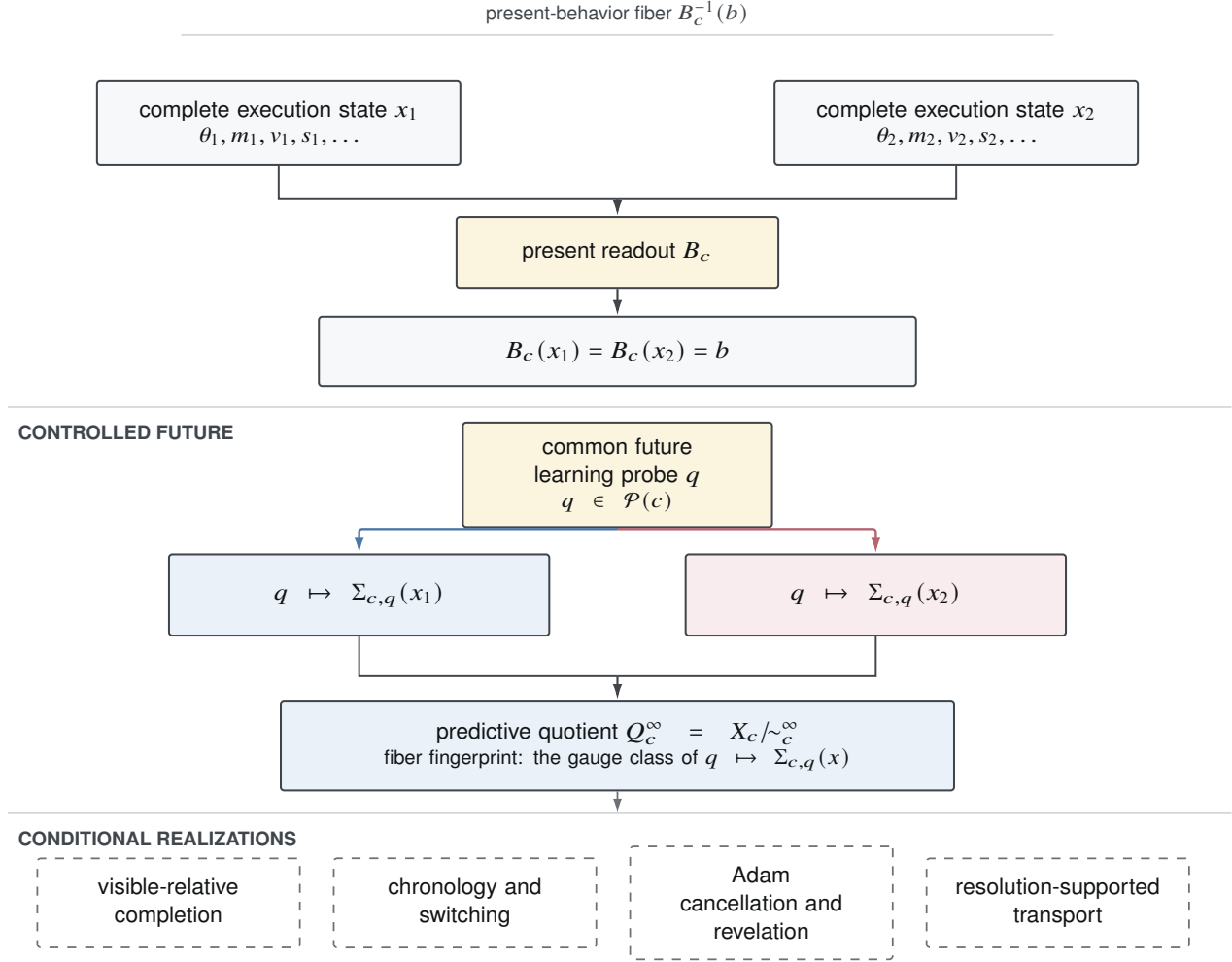
\begin{figure}[tbp]
    \centering
    \resizebox{0.98\linewidth}{!}{\begin{tikzpicture}[x=1cm,y=1cm]
  \node[ffpanel] at (0.0,8.75) {PRESENT EQUIVALENCE};
  \node[fflabel,text=FFGrayDark] at (7.20,8.18)
    {present-behavior fiber $B_c^{-1}(b)$};
  \draw[ffrule] (2.05,7.92)--(12.35,7.92);

  \node[ffneutral,minimum width=4.25cm,minimum height=1.00cm,text width=3.75cm]
    (x1) at (3.20,6.88) {complete execution state $x_1$\\[-1pt]
      {\scriptsize $\theta_1,m_1,v_1,s_1,\ldots$}};
  \node[ffneutral,minimum width=4.25cm,minimum height=1.00cm,text width=3.75cm]
    (x2) at (11.20,6.88) {complete execution state $x_2$\\[-1pt]
      {\scriptsize $\theta_2,m_2,v_2,s_2,\ldots$}};

  \coordinate (presentjoin) at (7.20,5.98);
  \draw[ffline] (x1.south)--(3.20,5.98)--(presentjoin);
  \draw[ffline] (x2.south)--(11.20,5.98)--(presentjoin);
  \node[ffgold,minimum width=3.80cm,minimum height=0.84cm]
    (readout) at (7.20,5.35) {present readout $B_c$};
  \draw[ffarrow] (presentjoin)--(readout.north);
  \node[ffneutral,minimum width=7.05cm,minimum height=0.82cm]
    (match) at (7.20,4.18) {$B_c(x_1)=B_c(x_2)=b$};
  \draw[ffarrow] (readout)--(match);

  \draw[ffrule] (0.00,3.52)--(14.45,3.52);
  \node[ffpanel] at (0.0,3.22) {CONTROLLED FUTURE};
  \node[ffgold,minimum width=3.65cm,minimum height=0.94cm,text width=3.10cm]
    (probe) at (7.20,2.72) {common future\\learning probe $q$\\[-1pt]
      {\scriptsize $q\in\mathcal P(c)$}};
  \node[ffblue,minimum width=4.45cm,minimum height=0.96cm,text width=3.95cm]
    (r1) at (4.15,1.30) {$q\mapsto\Sigma_{c,q}(x_1)$};
  \node[ffred,minimum width=4.45cm,minimum height=0.96cm,text width=3.95cm]
    (r2) at (10.25,1.30) {$q\mapsto\Sigma_{c,q}(x_2)$};
  \coordinate (probesplit) at (7.20,2.08);
  \draw[ffline] (probe.south)--(probesplit);
  \draw[ffarrowblue] (probesplit)-|(r1.north);
  \draw[ffarrowred] (probesplit)-|(r2.north);

  \coordinate (responsejoin) at (7.20,0.34);
  \draw[ffline] (r1.south)--(4.15,0.34)--(responsejoin);
  \draw[ffline] (r2.south)--(10.25,0.34)--(responsejoin);
  \node[ffblue,minimum width=7.95cm,minimum height=1.06cm,text width=7.30cm]
    (quotient) at (7.20,-0.48) {predictive quotient $Q_c^\infty=X_c/\!\sim_c^\infty$\\[-1pt]
      {\scriptsize fiber fingerprint: the gauge class of $q\mapsto\Sigma_{c,q}(x)$}};
  \draw[ffarrow] (responsejoin)--(quotient.north);

  \draw[ffrule] (0.00,-1.28)--(14.45,-1.28);
  \node[ffpanel] at (0.0,-1.58) {CONDITIONAL REALIZATIONS};
  \node[ffdashframe,minimum width=3.10cm,minimum height=0.78cm,text width=2.60cm]
    at (1.90,-2.34) {visible-relative\\completion};
  \node[ffdashframe,minimum width=3.10cm,minimum height=0.78cm,text width=2.60cm]
    at (5.40,-2.34) {chronology and\\switching};
  \node[ffdashframe,minimum width=3.10cm,minimum height=0.90cm,text width=2.60cm]
    at (8.90,-2.34) {Adam\\cancellation and\\revelation};
  \node[ffdashframe,minimum width=3.10cm,minimum height=0.78cm,text width=2.60cm]
    at (12.40,-2.34) {resolution-supported\\transport};
  \draw[ffdasharrow] (quotient.south)--(7.20,-1.28);
\end{tikzpicture}}
    \caption{Finite controlled core and conditional realizations.}
    \label{fig:architecture}
\end{figure}

The primitive chain is
\begin{equation}
\text{complete execution state}
\longrightarrow \text{present-behavior fiber}
\longrightarrow \text{future-response law}
\longrightarrow \text{predictive quotient}.
\label{eq:architecture-chain}
\end{equation}
A fiber fingerprint is the remaining future-response structure inside one present-behavior fiber. The construction is finite-scale and set-level first. Metric, Hilbert, differential, switching, optimizer, and smooth-global structures enter only as explicit enrichments under additional assumptions. This ordering matters because the predictive quotient remains well defined when dimensions jump, protocols are irreversible, or no global manifold exists; \cref{fig:architecture} summarizes the resulting hierarchy.

\subsection{Contributions}
\label{sec:contributions}

The paper makes four coupled contributions.
\begin{enumerate}[label=\textbf{C\arabic*.}]
    \item \textbf{Predictive state.} We define operational fiber fingerprints on present-behavior equivalence classes and prove predictive congruence, quotient descent, Nerode-type minimality, finite-horizon typing, a canonical set-level global fiber, and an exact Transformer--LoRA--AdamW instantiation.

    \item \textbf{History-reachable completion.} In a declared finite-dimensional Hilbert realization, future response decomposes into visible, visible-mode-reuse, and irreducible-new sectors. The history-weighted obstruction
    \begin{equation}
        \Birr=R^{1/2}J^*SJR^{1/2}
    \end{equation}
    determines the optimal finite-rank irreducible completion on naturally occupied history support.

    \item \textbf{Chronology and optimizer mechanisms.} A common-affine memory model admits an exact graph-coboundary plus interaction decomposition; cycle projection removes task-potential memory. Under a separately declared crossing regime, the cycle response has conditional root-mean-square scale $\sqrt p\,\eta^{3/2}$ with a graph-cyclomatic prefactor and finite-grid corrections. For Adam, an exact moment section preserves the immediate adaptive field while common future gradients can locally de-cancel and reveal hidden moment differences.

    \item \textbf{Frozen empirical membership.} Preregistered Qwen and Mistral studies test the local action backbone, visible-relative completion, transport, switching, matched-section accessibility, and future revelation. Positive, negative, and mixed outcomes are retained without pooling rescue, threshold revision, or post-outcome replacement of the estimand.
\end{enumerate}

\subsection{Scope and organization}

The finite predictive quotient and its set-level global fiber are unconditional relative to a declared probe doctrine. Completion, Hodge, switching, Adam, and smooth-global statements require their own finite-dimensionality, regularity, accessibility, stochastic, or spectral assumptions. Experiments test whether concrete Transformer execution contexts exhibit those conditional structures; they are not premises of the core theory.

This distinction turns the negative studies into informative boundaries. An unresolved finite response frame does not negate set-level transport; a finite-grid exponent outside a narrow equivalence band does not negate a separately stated asymptotic class; a raw-coordinate conditioning failure is not a coordinate-free rank theorem; and bank-dependent revelation is compatible with positive future observability.

\Cref{sec:core} develops the finite controlled core. \Cref{sec:completion} gives visible-relative completion and the history-reachability bridge. \Cref{sec:chronology,sec:adam,sec:transport} treat chronology, optimizer revelation, and supported transport. \Cref{sec:experiments} reports the Transformer studies, \cref{sec:related} states the novelty boundary, and \cref{sec:limits,sec:conclusion} give limitations and conclusions. Proof details, reproducibility contracts, and assumption ledgers appear in the appendices.

\section{Finite controlled experiments and predictive quotients}
\label{sec:core}

\subsection{Primitive finite experiment}

Let $\cC$ be a small category whose objects are declared execution contexts and whose arrows are admissible finite controlled-learning protocols. A state functor $X:\cC\to\mathbf{Set}$ assigns a complete state space $X_c$ to each context and a transition map $T_a:X_c\to X_d$ to every arrow $a:c\to d$. Thus
\begin{equation}
T_{\mathrm{id}_c}=\Id_{X_c},\qquad T_{b\circ a}=T_b\circ T_a.
\label{eq:complete-composition}
\end{equation}
No topology, probability law, linear structure, or differentiability is included in this primitive definition.

A probe doctrine assigns a set $\cP(c)$ of finite future-learning probes to each context. For $q\in\cP(c)$, the response map is $\Sigma_{c,q}:X_c\to Y_q$. If $a:c\to d$ and $q\in\cP(d)$, the prefixed probe $a^*q\in\cP(c)$ is required to satisfy
\begin{equation}
\Sigma_{d,q}(T_a x)=\Sigma_{c,a^*q}(x).
\label{eq:prefix}
\end{equation}
A declared present readout $B_c:X_c\to\mathcal B_c$ defines current behavior.

\begin{definition}[Present-behavior fiber]
For $b\in\mathcal B_c$, the present-behavior fiber is
\begin{equation}
F_{c,b}:=B_c^{-1}(b).
\end{equation}
It is a set-level object and need not be a manifold or have constant local dimension.
\end{definition}

\begin{definition}[Operational fiber fingerprint]
For $x\in F_{c,b}$ and a declared probe family $A\subseteq\cP(c)$, define
\begin{equation}
\FF^{\op}_{c,A}(x):=\bigl(\Sigma_{c,q}(x)\bigr)_{q\in A},
\label{eq:operational-ff}
\end{equation}
considered up to the declared response gauge. A hidden learning-state distinction is a pair $x,y$ such that $B_c(x)=B_c(y)$ but $\FF^{\op}_{c,A}(x)\neq\FF^{\op}_{c,A}(y)$.
\end{definition}

The definition is operational: it neither postulates a latent coordinate nor identifies the physical dimension of a model implementation. ``Complete'' below always means complete relative to the declared probe doctrine.

\subsection{Complete and finite-horizon equivalence}

\begin{definition}[Predictive equivalence]
For $x,y\in X_c$,
\begin{align}
x\sim_c^\infty y
&\iff \Sigma_{c,q}(x)=\Sigma_{c,q}(y)\quad\forall q\in\cP(c),\\
x\sim_c^{\le H}y
&\iff \Sigma_{c,q}(x)=\Sigma_{c,q}(y)\quad\forall q\in\cP(c)\text{ with }h(q)\le H.
\end{align}
Write $Q_c^\infty=X_c/\!\sim_c^\infty$ and $Q_c^{\le H}=X_c/\!\sim_c^{\le H}$.
\end{definition}

\begin{theorem}[Predictive congruence and quotient descent]
\label{thm:quotient-descent}
Under \cref{eq:complete-composition,eq:prefix}, predictive equivalence is a congruence: if $x\sim_c^\infty y$ and $a:c\to d$, then $T_ax\sim_d^\infty T_ay$. Consequently there is a unique descended map
\begin{equation}
\tau_a^\infty:Q_c^\infty\to Q_d^\infty,\qquad [x]\mapsto[T_ax],
\end{equation}
and $Q^\infty:\cC\to\mathbf{Set}$ is a functor.
\end{theorem}

\begin{proof}
For each $q\in\cP(d)$, prefix compatibility gives
\[
\Sigma_{d,q}(T_ax)=\Sigma_{c,a^*q}(x)=\Sigma_{c,a^*q}(y)=\Sigma_{d,q}(T_ay).
\]
Identity and composition descend from the state functor.
\end{proof}

The following minimality statement is the controlled-learning analogue of Nerode's quotient principle for automata \citep{nerode1958linear}.

\begin{theorem}[Nerode-type minimality]
\label{thm:nerode}
Let $z_c:X_c\to Z_c$ be a representation through which every declared response factors and on which every controlled protocol descends. Then
\begin{equation}
z_c(x)=z_c(y)\implies x\sim_c^\infty y.
\end{equation}
Hence the quotient map $q_c:X_c\to Q_c^\infty$ factors uniquely through $z_c$ on $z_c(X_c)$. The complete predictive quotient is the coarsest recursively sufficient representation relative to the declared probe doctrine.
\end{theorem}

\begin{proof}
Equality of $z_c(x)$ and $z_c(y)$ implies equality of every response that factors through $z_c$, so $x\sim_c^\infty y$. The quotient universal property yields the unique factor map.
\end{proof}

The global set-level object is obtained by the standard category-of-elements (Grothendieck) construction \citep{maclane1998categories}.

\begin{corollary}[Canonical global finite-scale fiber]
\label{cor:global-opfib}
The category of elements
\begin{equation}
\int_{\cC}Q^\infty\longrightarrow\cC
\end{equation}
is a canonical discrete opfibration. It exists even when quotient cardinalities or local ranks jump, contexts are singular, protocols are irreversible, and no global coordinate chart exists.
\end{corollary}

\begin{remark}[Finite-horizon typing]
If $a$ consumes $\ell(a)$ primitive steps, the correctly typed finite-horizon map is
\begin{equation}
\tau_{a,H}:Q_c^{\le H+\ell(a)}\to Q_d^{\le H}.
\end{equation}
Using the same horizon on both sides without an explicit convention is generally incorrect.
\end{remark}

\subsection{Exact prefix-depth fibers}

A present-behavior match can be strengthened by matching the responses generated along a fixed initial control prefix. Let $\alpha=(a_1,\ldots,a_k)$ be a declared composable prefix and let
\begin{equation}
U_j^\alpha:X_c\to\mathcal A_j
\end{equation}
be the declared action readout after the first $j-1$ controls of $\alpha$ have been applied. For example, $U_j^\alpha(x)$ may be the parameter increment or adaptive-field vector produced by control $a_j$ from the prefix state $T_{a_{j-1}}\cdots T_{a_1}x$. Define
\begin{equation}
C_\alpha^{(k)}(x)=\bigl(B_c(x),U_1^\alpha(x),\ldots,U_k^\alpha(x)\bigr).
\end{equation}

\begin{definition}[Exact prefix-depth-$k$ fiber]
\begin{equation}
F_{x,\alpha}^{(k)}:=\bigl(C_\alpha^{(k)}\bigr)^{-1}\!\bigl(C_\alpha^{(k)}(x)\bigr).
\end{equation}
For a compatible extension of the same prefix, $F_{x,\alpha}^{(k+1)}\subseteq F_{x,\alpha}^{(k)}$.
\end{definition}

These fibers let an experiment match present behavior and a declared initial learning-action sequence before asking whether longer futures differ. They remain exact set-level objects; tangent-kernel formulas require separate constant-rank, stratification, or metric-differentiability assumptions \citep{lee2013smooth}.

\subsection{Transformer execution-state instantiation}

Transformer architectures \citep{vaswani2017attention}, low-rank adapters \citep{hu2022lora}, and AdamW \citep{loshchilov2019adamw} provide one concrete execution context for the finite controlled core.

\begin{definition}[Complete Transformer execution state]
For a frozen Transformer--LoRA--AdamW context, the mutable state includes all trainable or master parameters, first and second moments, optimizer step, scheduler, gradient-accumulation buffers, loss-scaling and skipped-step state, random streams, data order and cursor, and every mutable cache or backend variable capable of altering future execution. Architecture, frozen weights, tokenizer and data assets, quantization metadata, hyperparameters, control alphabet, and numerical backend contract are context data.
\end{definition}

\begin{theorem}[Concrete finite-scale instantiation]
\label{thm:transformer-instantiation}
If each primitive execution step is single-valued given the complete state and primitive control, then finite Transformer--LoRA--AdamW training is a model of the controlled finite-scale axioms. It therefore possesses declared-probe predictive quotients, history-natural endpoints, directed quotient transport, and a global set-level predictive fiber without assuming smoothness or low rank.
\end{theorem}

\begin{proof}
Primitive steps are functions and protocols are finite compositions. Current behavior is a declared readout; a future probe is a finite suffix plus a readout schedule. Prefix compatibility is associativity of execution.
\end{proof}

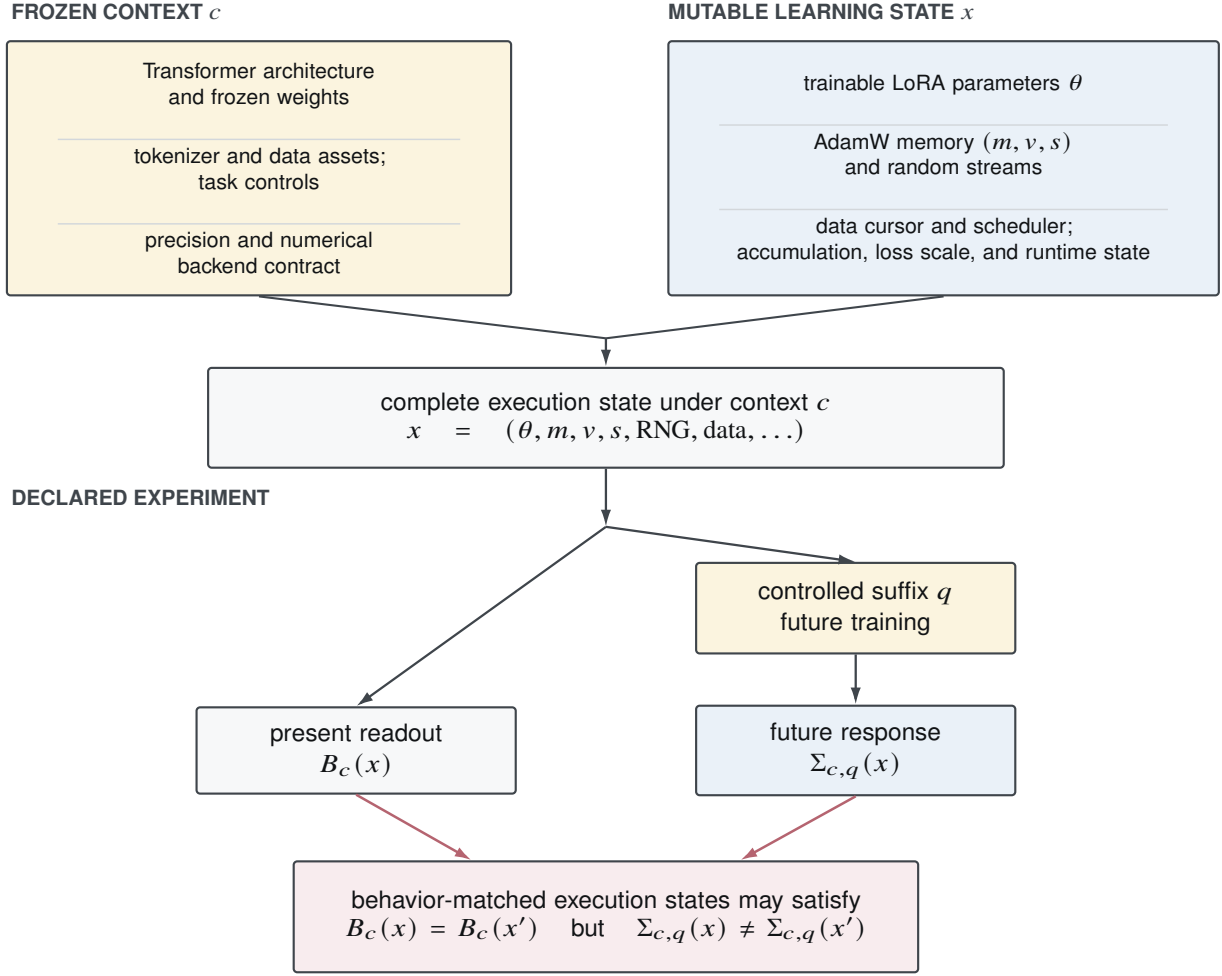
\begin{figure}[tbp]
    \centering
    \resizebox{0.98\linewidth}{!}{\begin{tikzpicture}[x=1cm,y=1cm]
  \node[ffpanel] at (0.0,9.55) {COMPLETE TRANSFORMER--LORA--ADAMW EXECUTION STATE};

  \node[ffpanel] at (0.45,8.88) {FROZEN CONTEXT $c$};
  \node[ffgold,font=\sffamily\scriptsize,minimum width=5.45cm,minimum height=2.75cm,text width=4.85cm]
    (context) at (3.25,7.18) {Transformer architecture\\and frozen weights\\[3pt]
      {\color{FFRule}\rule{4.35cm}{0.45pt}}\\[-1pt]
      tokenizer and data assets;\\task controls\\[3pt]
      {\color{FFRule}\rule{4.35cm}{0.45pt}}\\[-1pt]
      precision and numerical\\backend contract};

  \node[ffpanel] at (7.55,8.88) {MUTABLE LEARNING STATE $x$};
  \node[ffblue,font=\sffamily\scriptsize,minimum width=5.95cm,minimum height=2.75cm,text width=5.35cm]
    (mutable) at (10.65,7.18) {trainable LoRA parameters $\theta$\\[3pt]
      {\color{FFRule}\rule{4.85cm}{0.45pt}}\\[-1pt]
      AdamW memory $(m,v,s)$\\and random streams\\[3pt]
      {\color{FFRule}\rule{4.85cm}{0.45pt}}\\[-1pt]
      data cursor and scheduler;\\accumulation, loss scale, and runtime state};

  \coordinate (statejoin) at (7.00,5.34);
  \draw[ffline] (context.south)--(statejoin);
  \draw[ffline] (mutable.south)--(statejoin);
  \node[ffneutral,minimum width=8.60cm,minimum height=1.08cm,text width=7.95cm]
    (complete) at (7.00,4.48) {complete execution state under context $c$\\[-1pt]
      $x=(\theta,m,v,s,\mathrm{RNG},\mathrm{data},\ldots)$};
  \draw[ffarrow] (statejoin)--(complete.north);

  \node[ffpanel] at (0.45,3.62) {DECLARED EXPERIMENT};
  \coordinate (experimentsplit) at (7.00,3.30);
  \draw[ffarrow] (complete.south)--(experimentsplit);

  \node[ffgold,minimum width=3.45cm,minimum height=0.98cm,text width=2.90cm]
    (suffix) at (9.70,2.42) {controlled suffix $q$\\future training};
  \node[ffneutral,minimum width=3.45cm,minimum height=0.92cm,text width=2.90cm]
    (present) at (4.30,0.88) {present readout\\$B_c(x)$};
  \node[ffblue,minimum width=3.45cm,minimum height=0.92cm,text width=2.90cm]
    (future) at (9.70,0.88) {future response\\$\Sigma_{c,q}(x)$};

  \draw[ffarrow] (experimentsplit)--(present.north);
  \draw[ffarrow] (experimentsplit)--(suffix.north);
  \draw[ffarrow] (suffix.south)--(future.north);

  \node[ffred,minimum width=6.75cm,minimum height=1.20cm,text width=6.15cm]
    (claim) at (7.00,-0.92) {behavior-matched execution states may satisfy\\[-1pt]
      $B_c(x)=B_c(x')$ \quad but \quad $\Sigma_{c,q}(x)\neq\Sigma_{c,q}(x')$};
  \coordinate (claimleft) at ($(claim.north)+(-1.45cm,0)$);
  \coordinate (claimright) at ($(claim.north)+(1.45cm,0)$);
  \draw[ffarrowred] (present.south)--(claimleft);
  \draw[ffarrowred] (future.south)--(claimright);
\end{tikzpicture}}
    \caption{Complete Transformer--LoRA--AdamW execution state.}
    \label{fig:transformer-instantiation}
\end{figure}

The concrete split between context, mutable state, and declared readouts is summarized in \cref{fig:transformer-instantiation}.

\FloatBarrier

\begin{counterexample}[Physical dimension is not identifiable]
Given any controlled learner, take its direct product with an arbitrary auxiliary controlled system ignored by all declared responses. The operational fingerprint and predictive quotient are unchanged while implementation-state dimension can increase arbitrarily. Therefore quotient rank, bridge rank, and completion rank are operational dimensions, not physical neural-state dimensions.
\end{counterexample}

\begin{remark}[Approximate matching]
A tolerance relation such as $d(B_c(x),B_c(y))\le\epsilon$ is generally not transitive and need not be preserved by controlled transitions. Approximate matching therefore belongs to a separate measurement and stability layer. The exact quotient statements in this section do not silently replace equality by tolerance matching.
\end{remark}

\section{Visible-relative predictive completion}
\label{sec:completion}

The quotient theory is finite-scale and representation-free. This section introduces a conditional finite-dimensional Hilbert realization used to quantify which future distinctions are explained by a declared visible action backbone and which require additional predictive modes.

\subsection{Stable finite-dimensional realization}

Fix an anchor, a horizon $H$, and finite-dimensional Hilbert spaces of history coordinates $\cH$, intervention coordinates $\cU$, and cumulative response coordinates $\cY_{\le H}$. Let
\begin{equation}
J:\cH\to\cU,
\qquad
K_{\le H}:\cU\to\cY_{\le H}
\end{equation}
be local linear response operators in declared metrics. The finite-dimensional assumption is the main-text realization used for Moore--Penrose pseudoinverses \citep{penrose1955generalized}, Hilbert--Schmidt norms, traces, and spectral tails. Infinite-dimensional variants require closed-range and trace-class hypotheses, as summarized in \cref{sec:limits}.

These operators are not primitive. When a differential limit is unresolved, the finite-scale object remains the centered response contrast of a behavior-matched pair,
\begin{equation}
Z_{\epsilon,H}=\frac{\Sigma_{\le H}(x^+)-\Sigma_{\le H}(x^-)}{2}.
\end{equation}
The formulas below apply when a family of such contrasts admits a stable linear realization with fixed metrics and separated ranges.

\subsection{Visible, reuse, and irreducible-new sectors}

Let $P=P^*=P^2$ be a declared visible input projector on $\cU$ and let $Q=I-P$. Write $K=K_{\le H}$ and define the visible output projector
\begin{equation}
\Pi=(KP)(KP)^\dagger,
\label{eq:Pi-visible}
\end{equation}
which is the orthogonal projector onto $\Ran(KP)$. Then
\begin{equation}
K=K_{\mathrm{vis}}+K_{\mathrm{reuse}}+K_{\mathrm{new}},
\label{eq:vrpc-decomp}
\end{equation}
where
\begin{equation}
K_{\mathrm{vis}}=KP,
\qquad
K_{\mathrm{reuse}}=\Pi KQ,
\qquad
K_{\mathrm{new}}=(I-\Pi)KQ.
\label{eq:vrpc-terms}
\end{equation}
The first term is response generated by declared visible input directions; the second is input-invisible response that reuses the visible output range; the third occupies an output direction orthogonal to that range.

\begin{theorem}[Canonical relative decomposition]
\label{thm:relative-decomp}
For fixed $(K,P)$ and fixed Hilbert metrics, the three operators in \cref{eq:vrpc-terms} are pairwise Hilbert--Schmidt orthogonal and uniquely determined by the declared visible input and visible output ranges.
\end{theorem}

\begin{proof}
The sum follows from $P+Q=I$ and $\Pi+(I-\Pi)=I$. In finite dimension,
\[
\langle KP,\Pi KQ\rangle_{\HS}
=\tr(PK^*\Pi KQ)=\tr(QPK^*\Pi K)=0.
\]
The new component has output range orthogonal to $\Ran(KP)$ and therefore to both preceding terms. Uniqueness follows from the fixed orthogonal range classes.
\end{proof}

Write the Gramian $\Gamma=K^*K$ in the $P\oplus Q$ decomposition,
\begin{equation}
\Gamma=\begin{bmatrix}A&C\\ C^*&D\end{bmatrix},
\qquad
A=PK^*KP,\quad C=PK^*KQ,\quad D=QK^*KQ.
\end{equation}
Following the shorted-operator and range-factorization tradition \citep{anderson1975shorted,douglas1966majorization}, define the shorted residual, extended by zero on $P\cU$, by
\begin{equation}
S:=K_{\mathrm{new}}^*K_{\mathrm{new}}
=QK^*(I-\Pi)KQ
=D-C^*A^\dagger C\succeq0.
\label{eq:shorted}
\end{equation}

\begin{theorem}[Relative factorization obstruction]
\label{thm:factor-obstruction}
In the finite-dimensional realization,
\begin{equation}
\inf_{T:Q\cU\to P\cU}\|KQ-KPT\|_{\HS}^2
=\|K_{\mathrm{new}}\|_{\HS}^2
=\tr S.
\label{eq:factor-error}
\end{equation}
Consequently, $S=0$ if and only if $\Ran(KQ)\subseteq\Ran(KP)$.
\end{theorem}

\begin{proof}
For each column of $KQ$, the nearest vector in $\Ran(KP)$ is its orthogonal projection $\Pi KQ$. A linear preimage is supplied by $(KP)^\dagger\Pi KQ$. Summing squared residuals yields \cref{eq:factor-error}; the block identity follows from \cref{eq:Pi-visible}.
\end{proof}

\begin{theorem}[Minimal irreducible completion]
\label{thm:min-completion}
Allow an additional operator $R_m:Q\cU\to\cY_{\le H}$ with $\rank R_m\le m$ and $\Ran(R_m)\perp\Ran(KP)$. Then
\begin{equation}
\inf_{T,\,\rank R_m\le m}
\|KQ-KPT-R_m\|_{\HS}^2
=
\sum_{j>m}\lambda_j(S),
\label{eq:eckart-completion}
\end{equation}
where the eigenvalues are ordered decreasingly. Hence $\rank S$ is the finite-horizon irreducible completion dimension relative to the declared realization, not the physical state dimension.
\end{theorem}

\begin{proof}
After optimal visible projection the residual is $K_{\mathrm{new}}$. The Eckart--Young theorem \citep{eckart1936approximation} gives the optimal rank-$m$ approximation error, whose squared singular values are the eigenvalues of $S$.
\end{proof}

\begin{figure}[t]
    \centering
    \resizebox{0.98\linewidth}{!}{\begin{tikzpicture}[x=1cm,y=1cm]
  \node[ffpanel] at (0.0,9.55) {HISTORY-SUPPORTED RESPONSE COMPLETION};

  \node[ffgold,minimum width=4.20cm,minimum height=0.94cm,text width=3.65cm]
    (history) at (6.90,8.45) {natural-history direction $R^{1/2}h$};
  \node[ffneutral,minimum width=3.55cm,minimum height=0.90cm,text width=3.00cm]
    (intervention) at (6.90,6.95) {intervention space $\mathcal U$};
  \draw[ffarrowgold] (history.south)--node[right,fflabel]{$J$}(intervention.north);

  \node[ffblue,minimum width=3.25cm,minimum height=0.90cm,text width=2.70cm]
    (vis) at (3.35,5.35) {visible input $P\mathcal U$};
  \node[ffneutral,minimum width=3.25cm,minimum height=0.90cm,text width=2.70cm]
    (hid) at (9.00,5.35) {hidden input $Q\mathcal U$};
  \draw[ffarrowblue] (intervention.south)--(vis.north);
  \draw[ffarrow] (intervention.south)--(hid.north);

  \node[ffblue,minimum width=3.40cm,minimum height=0.92cm,text width=2.86cm]
    (kvis) at (3.35,3.55) {visible response\\$K_{\rm vis}=KP$};
  \node[ffgold,minimum width=3.55cm,minimum height=0.92cm,text width=3.00cm]
    (reuse) at (7.10,3.55) {visible-range reuse\\$K_{\rm reuse}=\Pi KQ$};
  \node[ffred,minimum width=3.95cm,minimum height=0.98cm,text width=3.35cm]
    (new) at (11.00,3.55) {irreducible-new response\\$K_{\rm new}=(I-\Pi)KQ$};

  \draw[ffarrowblue] (vis.south)--(kvis.north);
  \draw[ffarrowgold] (hid.south)--(reuse.north);
  \draw[ffarrowred] (hid.south)--(new.north);

  \node[ffneutral,minimum width=5.85cm,minimum height=1.48cm,text width=5.23cm]
    (bridge) at (11.00,0.82) {{\scriptsize\bfseries HISTORY-WEIGHTED OBSTRUCTION}\\[-1pt]
      $\displaystyle \mathfrak B^{\mathrm{irr}}
      =(K_{\rm new}JR^{1/2})^*(K_{\rm new}JR^{1/2})$\\[-1pt]
      {\scriptsize naturally occupied response beyond the visible backbone}};
  \draw[ffarrowred] (new.south)--(bridge.north);
\end{tikzpicture}}
    \caption{Visible-relative completion on history-supported directions.}
    \label{fig:decomposition}
\end{figure}
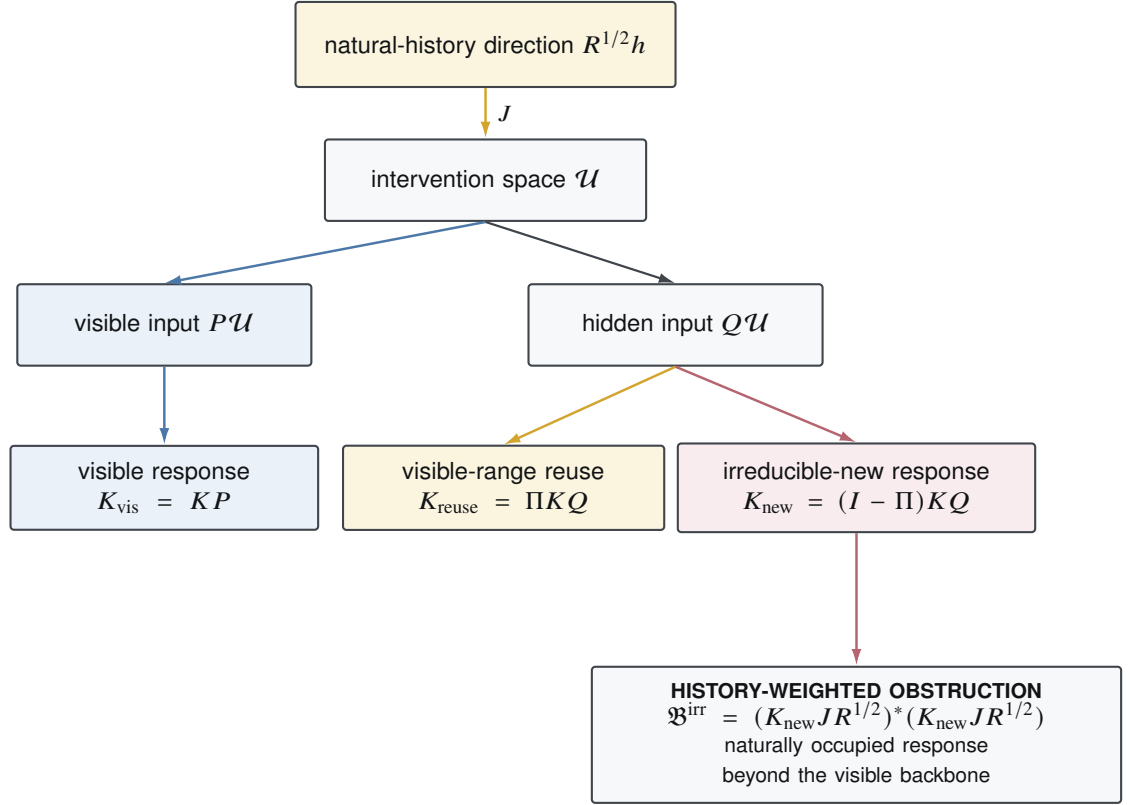

\Cref{fig:decomposition} summarizes the visible/reuse/new split and the subsequent history weighting.

\subsection{History-reachability bridge}

Let $R\succeq0$ be the covariance of naturally generated history coordinates. Set
\begin{equation}
N:=(I-\Pi)KQ,
\qquad
\Birr:=R^{1/2}J^*SJ R^{1/2}
=(NJR^{1/2})^*(NJR^{1/2}).
\label{eq:bridge}
\end{equation}

\begin{theorem}[History-weighted factorization bridge]
\label{thm:history-bridge}
For linear maps $L:\cH\to P\cU$,
\begin{equation}
\inf_L
\|(KQJ-KPL)R^{1/2}\|_{\HS}^2
=\tr\Birr.
\label{eq:history-factor}
\end{equation}
Moreover, $\Birr=0$ exactly when every naturally occupied response in $KQJ\Ran R^{1/2}$ lies in the visible output range $\Ran(KP)$.
\end{theorem}

\begin{proof}
Apply \cref{thm:factor-obstruction} to the covariance-weighted operator $KQJR^{1/2}$. The orthogonal residual is $NJR^{1/2}$ and its squared Hilbert--Schmidt norm is $\tr\Birr$.
\end{proof}

\begin{corollary}[Reachable minimal completion]
The optimal error after adding $m$ genuinely new reachable modes is
\begin{equation}
\sum_{j>m}\lambda_j(\Birr).
\end{equation}
The covariance-supported quotient by $\Ker(NJR^{1/2})$ is the minimal reachable irreducible predictive completion.
\end{corollary}

The bridge prevents two overclaims. First, $S\neq0$ does not imply $\Birr\neq0$: a geometric response direction may exist but never be generated by natural histories. Second, low completion rank does not identify neural-state dimension; it identifies only the number of response modes required on the occupied support.

\subsection{Cumulative horizon filtration}

For a nested response family $K_{\le H}u=(K_1u,\ldots,K_Hu)$ with a compatible direct-sum metric and fixed $(P,J,R)$, let $S_{\le H}$ and $\Birr_{\le H}$ be the corresponding operators.

\begin{theorem}[Monotone irreducible filtration]
\label{thm:monotone-filtration}
\begin{equation}
0\preceq S_{\le1}\preceq S_{\le2}\preceq\cdots,
\qquad
0\preceq \Birr_{\le1}\preceq\Birr_{\le2}\preceq\cdots.
\end{equation}
Therefore kernels can only shrink and finite-dimensional completion ranks cannot decrease with cumulative horizon.
\end{theorem}

\begin{proof}
For $u\in\cU$,
\[
\langle u,S_{\le H}u\rangle
=
\inf_{v\in P\cU}\sum_{h\le H}\|K_hQu-K_hPv\|^2.
\]
Adding a nonnegative horizon term cannot reduce the infimum. Congruence by $JR^{1/2}$ gives the bridge inequality.
\end{proof}

\subsection{Complete prediction versus irreducible novelty}

Let $E$ be a finite-dimensional reachable local chart and let $D:E\to Y$ be the differential of the complete declared response law. Let $V\subseteq Y$ be the visible/reusable response subspace. Define
\begin{equation}
N_{\mathrm{pred}}=\Ker D,
\qquad
N_{\mathrm{irr}}=\Ker[(I-\Pi_V)D]=D^{-1}(V).
\end{equation}

\begin{theorem}[Predictive exact sequence]
\label{thm:predictive-exact-sequence}
There is a canonical short exact sequence
\begin{equation}
0\longrightarrow N_{\mathrm{irr}}/N_{\mathrm{pred}}
\longrightarrow E/N_{\mathrm{pred}}
\longrightarrow E/N_{\mathrm{irr}}
\longrightarrow0.
\label{eq:predictive-exact}
\end{equation}
The middle term is the complete reachable predictive tangent, the right term is the irreducible bridge quotient, and the kernel consists of future distinctions expressed entirely through visible-mode reuse.
\end{theorem}

\begin{proof}
The map is induced by $N_{\mathrm{pred}}\subseteq N_{\mathrm{irr}}$. Its kernel is $N_{\mathrm{irr}}/N_{\mathrm{pred}}$, and it is surjective.
\end{proof}

\subsection{Gauge and stability boundary}

Orthogonality, spectra, pseudoinverses, and shorted operators are canonical only relative to declared metrics and stable ranges. Under unitary gauge changes, nonzero spectra are invariant. Under general coordinates, the metrics must be transported as well. If a singular gap closes, pseudoinverses and projectors need not converge even when finite responses do. The finite response law is therefore primary; the operator layer is a stable realization rather than an ontology.

\section{History generation, chronology, and switching}
\label{sec:chronology}

The response geometry of \cref{sec:completion} describes what can be revealed. A separate module describes what training histories generate and how task order organizes the generated directions.

\subsection{History-natural occupancy}

Let $\mathsf H_c$ be rooted histories ending at context $c$, let $e_c:\mathsf H_c\to X_c$ be the endpoint map, and let $\nu_c$ be a declared history law. Its pushforward $\mu_c=(e_c)_\#\nu_c$ is the reachable state law. Under a local Hilbert chart, the centered history coordinate has covariance $R_c\succeq0$. History naturality requires
\begin{equation}
T_a\circ e_c=e_d\circ\sigma_a
\end{equation}
for history continuation $\sigma_a$. On regular quotient charts, differentiation gives
\begin{equation}
L_aJ_c=J_dS_a,
\end{equation}
where $S_a$ transports history coordinates and $L_a$ transports predictive coordinates. The set-level commuting square is primary; the differential relation requires the stated local charts and differentiability.

\subsection{Task-graph chronology and Hodge sectors}

Let $G=(V,E)$ be a finite oriented graph whose vertices index task types. An oriented edge $e=(a,b)$ is taken from task $a$ to task $b$. Let $d\in\ell^2(E;\cH)$ be a hidden-state-valued edge field constructed from ordered task pairs. With incidence operator $B_G$, define the coboundary
\begin{equation}
\delta:=B_G^*\otimes I_{\cH},
\qquad
(\delta\psi)_{ab}=\psi_b-\psi_a.
\label{eq:coboundary-convention}
\end{equation}
Hilbert-valued graph Hodge theory \citep{jiang2011hodge} gives
\begin{equation}
\ell^2(E;\cH)=\Ran(\delta)\oplus\Ker(B_G\otimes I_{\cH}).
\label{eq:hodge}
\end{equation}
Write $d=d_{\mathrm{ex}}+d_{\mathrm{cyc}}$ and let $P_{\mathrm{cyc}}$ denote the orthogonal projector onto the cycle sector. The decomposition itself is standard; the mechanism-specific result is the interpretation of the two sectors under optimizer memory.

\subsection{Common-affine memory isolates interaction}

Suppose a hidden memory channel obeys
\begin{equation}
z(\Phi_a x)=Az(x)+U_a(x)
\label{eq:common-affine}
\end{equation}
with one common bounded operator $A$. Define the half order defect on the oriented edge $(a,b)$ by
\begin{equation}
d_{ab}(x)=\frac12\bigl(z(\Phi_b\Phi_a x)-z(\Phi_a\Phi_b x)\bigr)
\end{equation}
and the task-interaction increment by
\begin{equation}
\Delta_aU_b(x)=U_b(\Phi_a x)-U_b(x).
\end{equation}

\begin{theorem}[Coboundary--interaction decomposition]
\label{thm:coboundary-interaction}
Under \cref{eq:common-affine},
\begin{equation}
d=\delta\psi+\omega,
\qquad
\psi_a=\frac12(I-A)U_a,
\qquad
\omega_{ab}=\frac12(\Delta_aU_b-\Delta_bU_a).
\label{eq:coboundary-interaction}
\end{equation}
Consequently,
\begin{equation}
P_{\mathrm{cyc}}d=P_{\mathrm{cyc}}\omega.
\label{eq:cycle-isolation}
\end{equation}
Cycle projection removes all common-affine single-task potential memory and retains the state-dependent interaction obstruction.
\end{theorem}

\begin{proof}
Expanding \cref{eq:common-affine} twice gives
\[
z(\Phi_b\Phi_a x)=A^2z(x)+AU_a(x)+U_b(x)+\Delta_aU_b(x)
\]
and the analogous expression with $a,b$ exchanged. Taking the half difference yields
\[
\frac12(I-A)(U_b-U_a)+\frac12(\Delta_aU_b-\Delta_bU_a),
\]
whose first term is $(\delta\psi)_{ab}$ under \cref{eq:coboundary-convention}. The cycle projector annihilates every coboundary.
\end{proof}

For task-dependent operators $A_a$, an additional contraction/commutator obstruction remains. Thus \cref{eq:cycle-isolation} is exact only in the common-operator specialization or after the heterogeneous term is explicitly retained.

\subsection{Smooth and switching chronology}

For smooth update fields, fix the convention
\begin{equation}
T_a^\eta(x)=x+\eta F_a(x),
\qquad
[F_a,F_b]:=DF_b\,F_a-DF_a\,F_b.
\label{eq:bracket-convention}
\end{equation}
A Taylor expansion then gives
\begin{equation}
T_b^\eta T_a^\eta-T_a^\eta T_b^\eta
=\eta^2[F_a,F_b]+O(\eta^3).
\label{eq:smooth-bracket}
\end{equation}
This is the established smooth-order effect, recently used to analyze learning-order geometry \citep{rukhovich2025commute,sweeney2026geometry}. In a piecewise-smooth system, boundary crossings add jump terms that need not share the same scale.

\begin{assumption}[Regular graph-valued crossing regime]
\label{ass:regular-crossing}
For edge crossing fields $C_e^\eta\in\cH$ on a fixed graph, as $p\to\infty$, $\eta\downarrow0$, and $p\eta\to\infty$,
\begin{equation}
\E\langle C_e^\eta,C_f^\eta\rangle
=\delta_{ef}\,q_e\,p\eta^3+o(p\eta^3)
\label{eq:cross-cov}
\end{equation}
uniformly in $e,f$, and the sum of all off-diagonal covariance remainders is $o(p\eta^3)$.
\end{assumption}

The assumption packages a regular crossing count of order $p\eta$, post-update centered marks of order $\eta$, finite second moments, and sufficiently weak aggregate dependence. It is not implied merely by the presence of ReLU units.

\begin{theorem}[Graph-projected $3/2$ law]
\label{thm:hodge-three-halves}
Under \cref{ass:regular-crossing},
\begin{equation}
\E\|(P_{\mathrm{cyc}}\otimes I)C^\eta\|^2
=p\eta^3\sum_{e\in E}(P_{\mathrm{cyc}})_{ee}q_e+o(p\eta^3).
\label{eq:hodge-three-halves}
\end{equation}
If $q_e=q$ for every edge, then
\begin{equation}
\E\|C^\eta_{\mathrm{cyc}}\|^2
=q\,b_1(G)\,p\eta^3+o(p\eta^3),
\qquad
\|C^\eta_{\mathrm{cyc}}\|_{\mathrm{RMS}}
\sim\sqrt{q\,b_1(G)p}\,\eta^{3/2}.
\end{equation}
\end{theorem}

\begin{proof}
Since $P_{\mathrm{cyc}}$ is an orthogonal projector,
\[
\E\|P_{\mathrm{cyc}}C\|^2
=\sum_{e,f}(P_{\mathrm{cyc}})_{ef}\E\langle C_e,C_f\rangle.
\]
Insert \cref{eq:cross-cov}. In the homogeneous case, $\tr P_{\mathrm{cyc}}=\rank P_{\mathrm{cyc}}=b_1(G)$.
\end{proof}

More generally, if the event count scales as $\eta^\alpha$, the event mark as $\eta^r$, and aggregation as the $\theta$th power of the event count, then the norm exponent is $r+\alpha\theta$. The value $3/2$ is one conditional universality class, not a universal consequence of a named architecture. \Cref{fig:chronology-hodge} summarizes the interaction isolation and the inputs to the regular switching class.

\begin{figure}[tbp]
    \centering
    \resizebox{0.98\linewidth}{!}{\begin{tikzpicture}[x=1cm,y=1cm]
  \node[ffpanel] at (0.0,10.20) {(a) CHRONOLOGY FIELD AND HODGE ISOLATION};

  \node[circle,ffblue,minimum size=0.82cm,inner sep=0pt] (A) at (1.95,8.72) {$A$};
  \node[circle,ffgold,minimum size=0.82cm,inner sep=0pt] (B) at (0.82,6.98) {$B$};
  \node[circle,ffred,minimum size=0.82cm,inner sep=0pt] (C) at (3.08,6.98) {$C$};
  \draw[ffarrow] (A)--node[left,fflabel]{$d_{AB}$}(B);
  \draw[ffarrow] (B)--node[below,fflabel]{$d_{BC}$}(C);
  \draw[ffarrow] (C)--node[right,fflabel]{$d_{CA}$}(A);

  \node[ffneutral,minimum width=4.15cm,minimum height=0.98cm,text width=3.55cm]
    (decomp) at (7.15,6.98) {hidden-state edge field\\$d=\delta\psi+\omega$};
  \draw[ffarrow] (C.east)--node[above=1.2mm,fflabel,fill=white,inner sep=1.2pt]{assemble $d$}(decomp.west);

  \node[ffblue,minimum width=3.25cm,minimum height=1.02cm,text width=2.70cm]
    (exact) at (5.05,5.20) {exact sector\\$d_{\rm ex}=\delta\psi$};
  \node[ffred,minimum width=3.25cm,minimum height=1.02cm,text width=2.70cm]
    (cycle) at (9.25,5.20) {cycle sector\\$d_{\rm cyc}=P_{\rm cyc}\omega$};
  \draw[ffarrowblue] (decomp.south)--(exact.north);
  \draw[ffarrowred] (decomp.south)--(cycle.north);
  \node[ffsubtle] at (5.05,4.42) {task-potential memory};
  \node[ffsubtle] at (9.25,4.42) {state-dependent interaction};

  \draw[ffrule] (0.0,3.93)--(14.35,3.93);

  \node[ffpanel] at (0.0,3.57) {(b) REGULAR CYCLE-PROJECTED SWITCHING CLASS};
  \node[ffgold,minimum width=3.15cm,minimum height=0.94cm,text width=2.58cm]
    (count) at (2.25,2.50) {crossing count\\$N\asymp p\eta$};
  \node[ffblue,minimum width=3.15cm,minimum height=0.94cm,text width=2.58cm]
    (mark) at (7.15,2.50) {centered mark\\$\|M_e\|\asymp\eta$};
  \node[ffneutral,minimum width=3.45cm,minimum height=0.94cm,text width=2.88cm]
    (depend) at (12.05,2.50) {weak aggregate dependence};

  \node[ffneutral,minimum width=5.10cm,minimum height=0.96cm,text width=4.52cm]
    (aggregate) at (7.15,1.10) {square-root aggregation\\after cycle projection};
  \draw[ffarrow] (count.south)--([xshift=-1.55cm]aggregate.north);
  \draw[ffarrow] (mark.south)--(aggregate.north);
  \draw[ffarrow] (depend.south)--([xshift=1.55cm]aggregate.north);

  \node[ffred,minimum width=6.65cm,minimum height=1.12cm,text width=6.05cm]
    (law) at (7.15,-0.43) {$\displaystyle
      \|d_{\rm cyc}^{\rm cross}\|_{\rm RMS}
      \asymp \sqrt{q\,b_1(G)\,p}\,\eta^{3/2}$};
  \draw[ffarrowred] (aggregate.south)--(law.north);
  \node[ffsubtle,text width=9.2cm] at (7.15,-1.40)
    {centering $\cdot$ finite variance $\cdot$ weak dependence $\cdot$ declared joint limit $p\eta\to\infty$};
\end{tikzpicture}}
    \caption{Chronology interaction and the regular switching class.}
    \label{fig:chronology-hodge}
\end{figure}

\subsection{Second-order finite-scale correction}

A leading asymptotic law need not produce its limiting exponent on a finite grid. We state the correction under an explicit concentration condition.

\begin{assumption}[Second-order energy and rare-event concentration]
\label{ass:second-order-concentration}
For a graph-projected cycle response $R_{p,\eta}$, suppose
\begin{equation}
\E R_{p,\eta}^2
=p\eta^3\bigl(\kappa_0+\kappa_1\eta+\kappa_2\eta^2+o(\eta^2)\bigr),
\qquad \kappa_0>0,
\label{eq:second-order-energy}
\end{equation}
and
\begin{equation}
\frac{R_{p,\eta}}{\sqrt{\E R_{p,\eta}^2}}
=1+O_p\bigl((p\eta)^{-1/2}\bigr).
\label{eq:rare-event-concentration}
\end{equation}
\end{assumption}

Bounded dependency graphs with suitable fourth-moment control are one sufficient route to \cref{eq:rare-event-concentration}; the paper uses the concentration statement itself as the named assumption.

\begin{proposition}[Second-order effective exponent]
\label{prop:second-order-law}
Under \cref{ass:second-order-concentration},
\begin{equation}
R_{p,\eta}
=\sqrt p\,\eta^{3/2}
\sqrt{\kappa_0+\kappa_1\eta+\kappa_2\eta^2+o(\eta^2)}
\left[1+O_p\bigl((p\eta)^{-1/2}\bigr)\right].
\label{eq:second-order-law}
\end{equation}
The local log-slope of the deterministic amplitude factor is
\begin{equation}
\beta_{\mathrm{eff}}(\eta)
=\frac32+\frac12
\frac{\kappa_1\eta+2\kappa_2\eta^2+o(\eta^2)}
{\kappa_0+\kappa_1\eta+\kappa_2\eta^2+o(\eta^2)}.
\label{eq:effective-exponent}
\end{equation}
\end{proposition}

\begin{proof}
Take the square root of \cref{eq:second-order-energy}, multiply by the relative concentration factor in \cref{eq:rare-event-concentration}, and differentiate the logarithm of the deterministic amplitude.
\end{proof}

The effective event count is $p\eta$, so the generic joint-limit fluctuation is $(p\eta)^{-1/2}$ rather than unconditionally $p^{-1/2}$.

\subsection{Generation and future observability}

Chronology generation and future observability are distinct. Let $O_{c,b,H}$ be a future-bank response operator acting on a reachable hidden direction $u$. The positive-semidefinite observability field is
\begin{equation}
W_{c,H}=\sum_b\pi_b O_{c,b,H}^*O_{c,b,H},
\qquad
\langle u,W_{c,H}u\rangle
=\sum_b\pi_b\|O_{c,b,H}u\|^2.
\label{eq:context-observability}
\end{equation}
Positive generation covariance can lie in $\Ker W_{c,H}$; conversely, a highly revealing direction may be rarely generated. Storage and revelation efficiency likewise factor, so weak visibility alone does not identify rapid dissipation. This separation is essential for interpreting low-gain or negative empirical results.

\section{Adam execution fibers, accessibility, and future revelation}
\label{sec:adam}

Adam and AdamW \citep{kingma2015adam,loshchilov2019adamw} provide an exact optimizer specialization in which current action equivalence can be constructed algebraically and future gradients can break that equivalence.

\subsection{Exact adaptive field and execution fiber}

For one coordinate at optimizer step $s$, let
\begin{equation}
\widehat m_s=\frac{m_s}{c_{1,s}},
\qquad
\widehat v_s=\frac{v_s}{c_{2,s}},
\qquad
\phi_s=\frac{\widehat m_s}{\sqrt{\widehat v_s}+\epsilon},
\end{equation}
where $c_{1,s}=1-\beta_1^s$ and $c_{2,s}=1-\beta_2^s$. With fixed parameter $\theta_s$, AdamW produces
\begin{equation}
\theta_{s+1}=(1-\eta_s\lambda)\theta_s-\eta_s\phi_s.
\end{equation}
An immediate-execution fiber is a level set of $\phi_s$ in $(m_s,v_s)$ space.

Set
\begin{equation}
r_s=\sqrt{v_s/c_{2,s}},
\qquad
D_s=r_s+\epsilon.
\end{equation}
On a branch where $e^\xi D_s-\epsilon\ge0$, define
\begin{equation}
m_s(\xi)=e^\xi m_s,
\qquad
v_s(\xi)=c_{2,s}[e^\xi D_s-\epsilon]^2.
\label{eq:matched-section}
\end{equation}

\begin{theorem}[Exact matched-section cancellation]
\label{thm:matched-section}
Along \cref{eq:matched-section},
\begin{equation}
\phi_s(\xi)=\phi_s(0)
\end{equation}
exactly. Thus different moment states can share the same current parameter, present behavior, and immediate Adam adaptive field.
\end{theorem}

\begin{proof}
The denominator becomes $\sqrt{v_s(\xi)/c_{2,s}}+\epsilon=e^\xi D_s$, while the numerator is multiplied by $e^\xi$. The factors cancel.
\end{proof}

\subsection{Moment reachability versus neural accessibility}

Over $K$ scalar gradient injections, write
\begin{equation}
M_K=\sum_{j=1}^K a_jg_j,
\qquad
V_K=\sum_{j=1}^K b_jg_j^2,
\end{equation}
with positive Adam weights $a_j,b_j$, and define
\begin{equation}
C_K=\sum_{j=1}^K\frac{a_j^2}{b_j}.
\end{equation}

\begin{theorem}[Exact scalar reachable moment region]
\label{thm:moment-epigraph}
For $K\ge2$, the reachable terminal injections are exactly the epigraph
\begin{equation}
V_K\ge\frac{M_K^2}{C_K}.
\label{eq:moment-epigraph}
\end{equation}
For $K=1$, equality is required.
\end{theorem}

\begin{proof}
Weighted Cauchy--Schwarz gives $M_K^2\le C_KV_K$. Equality is attained by the minimum-energy gradient history. For $K\ge2$, a nonzero component in the nullspace of the first-moment functional leaves $M_K$ unchanged and increases $V_K$ continuously, filling the epigraph.
\end{proof}

The theorem is optimizer-level. Real neural gradients are coupled across coordinates and constrained by the model and data.

\begin{theorem}[Local neural accessibility]
\label{thm:neural-accessibility}
Let $U$ and $M$ be finite-dimensional control and moment-coordinate spaces. Let $\Gamma_K:U\to G^K$ be a $C^1$ map from admissible controls to a $K$-step gradient sequence, let $\mathcal M_K:G^K\to M$ be the Adam moment map, and put
\begin{equation}
F_K=\mathcal M_K\circ\Gamma_K.
\end{equation}
If $DF_K(c_*)$ is surjective, then $F_K(U)$ contains a neighborhood of $F_K(c_*)$. Any matched-section point in that neighborhood is dynamically reachable by nearby controls.
\end{theorem}

\begin{proof}
This is the finite-dimensional submersion theorem \citep{lee2013smooth}.
\end{proof}

Surjectivity and stable numerical accessibility are distinct. Let $G_U,G_M$ be declared positive-definite metrics. Define
\begin{equation}
\widetilde J=G_M^{1/2}(DF_K)G_U^{-1/2},
\qquad
\sigma_{\min}^{G_U,G_M}(DF_K)=\sigma_{\min}(\widetilde J)
\label{eq:metric-jacobian}
\end{equation}
when $DF_K$ is surjective. These generalized singular values are invariant under coordinate changes when the metrics are transported; a raw Euclidean condition number is not.

\begin{proposition}[Metric-stable reachable radius]
\label{prop:stable-radius}
Let $F:B_U(0,r_0)\to M$ be $C^1$ between finite-dimensional metric spaces. Suppose $DF(0)$ is surjective,
\begin{equation}
\sigma_{\min}^{G_U,G_M}(DF(0))=\tau>0,
\end{equation}
and, for $u$ in the radius-$r_0$ ball,
\begin{equation}
\|DF(u)-DF(0)\|_{G_U\to G_M}\le L\|u\|_{G_U}.
\end{equation}
Then, for every $r\le\min\{r_0,\tau/(2L)\}$,
\begin{equation}
B_M\!\left(F(0),\frac{\tau r}{2}\right)
\subseteq F\bigl(B_U(0,r)\bigr).
\label{eq:stable-radius}
\end{equation}
\end{proposition}

This yields four distinct levels: algebraic reachability, local accessibility, metric-stable accessibility, and membership in a finite declared control panel.

\subsection{Future-gradient de-cancellation}

Prescribe exogenous common future gradients $g_1,\ldots,g_n$ shared by all points on the matched section. Let
\begin{align}
M_n&=\beta_1^n m_s,
& G_n^{(m)}&=(1-\beta_1)\sum_{j=1}^n\beta_1^{n-j}g_j,\\
A_n&=\beta_2^n c_{2,s},
& G_n^{(v)}&=(1-\beta_2)\sum_{j=1}^n\beta_2^{n-j}g_j^2.
\end{align}
At $\xi=0$, define
\begin{equation}
v_n^0=A_n(D_s-\epsilon)^2+G_n^{(v)},
\qquad
r_n=\sqrt{v_n^0/c_{2,s+n}},
\label{eq:future-root}
\end{equation}
and, whenever the denominators are nonzero,
\begin{equation}
\mu_n=\frac{M_n}{M_n+G_n^{(m)}},
\qquad
\nu_n=
\frac{A_n(D_s-\epsilon)D_s}
{c_{2,s+n}r_n(r_n+\epsilon)}.
\label{eq:memory-fractions}
\end{equation}

\begin{theorem}[First-order future-gradient revelation]
\label{thm:future-revelation}
Where \cref{eq:memory-fractions} is defined,
\begin{equation}
\left.\partial_\xi\phi_{s+n}(\xi)\right|_{\xi=0}
=\phi_{s+n}(0)(\mu_n-\nu_n).
\label{eq:revelation-derivative}
\end{equation}
The matched section is first-order locally resolved at horizon $n$ in a coordinate whenever the right-hand side is nonzero. At $n=0$, the derivative vanishes identically; when the fractions are defined, $\mu_0=\nu_0=1$.
\end{theorem}

\begin{proof}
The future first moment is affine in $e^\xi$, with logarithmic derivative $\mu_n$. Differentiating the future second moment and the square-root denominator yields $\nu_n$. Subtracting the two logarithmic derivatives gives \cref{eq:revelation-derivative}. The $n=0$ statement also follows directly from \cref{thm:matched-section}.
\end{proof}

Unequal decay rates alone are insufficient: if every future $g_j=0$ and $\epsilon=0$, numerator and square-root denominator retain the same $e^\xi$ scaling for all $n$.

\subsection{Revelation is not coherence}

For a reachable hidden direction $u$ and future bank $b$, let
\begin{equation}
\Delta_b=O_bu.
\end{equation}
The positive-semidefinite observability energy is
\begin{equation}
\sum_b\pi_b\|O_bu\|^2
=\left\langle u,\sum_b\pi_bO_b^*O_bu\right\rangle.
\end{equation}
Cross-bank coherence instead measures $\langle O_bu,O_{b'}u\rangle$.

\begin{proposition}[Revelation--coherence separation]
\label{prop:revelation-coherence}
Positive revelation by every bank does not imply positive cross-bank directional coherence. In particular, with $O_1=I$ and $O_2=R_\theta$ a planar rotation,
\begin{equation}
\|O_1u\|=\|O_2u\|=\|u\|>0,
\qquad
\frac{\langle O_1u,O_2u\rangle}{\|O_1u\|\|O_2u\|}=\cos\theta,
\end{equation}
which can take any value in $[-1,1]$.
\end{proposition}

A bank-invariant direction requires an additional homogeneous condition, for example $O_b|_S=\lambda_bO_0|_S$ with $\lambda_b>0$. The general theory therefore uses a context-indexed observability field rather than one universal resolving vector. \Cref{fig:adam-revelation} summarizes exact current cancellation and future-gradient de-cancellation.

\begin{figure}[tbp]
    \centering
    \resizebox{0.98\linewidth}{!}{\begin{tikzpicture}[x=1cm,y=1cm]
  \node[ffpanel] at (0.0,7.75) {(a) EXACT MATCHED SECTION};
  \node[ffpanel] at (6.85,7.75) {(b) FUTURE-GRADIENT DE-CANCELLATION};

  \draw[->,draw=FFGrayDark,line width=0.64pt] (0.75,1.55)--(0.75,6.72)
    node[above,fflabel]{$m$};
  \draw[->,draw=FFGrayDark,line width=0.64pt] (0.75,1.55)--(5.55,1.55)
    node[right,fflabel]{$\sqrt v$};
  \draw[draw=FFBlue,line width=1.15pt,domain=0.62:4.35,smooth,samples=90]
    plot (\x,{1.38+0.50*exp(0.32*\x)});
  \node[fflabel,text=FFBlue] at (3.50,5.75) {exact matched section};
  \filldraw[fill=FFBlue,draw=white,line width=0.45pt] (1.62,2.25) circle (2.45pt);
  \filldraw[fill=FFRed,draw=white,line width=0.45pt] (4.15,3.28) circle (2.45pt);
  \node[fflabel,anchor=east] at (1.48,2.22) {$x_-$};
  \node[fflabel,anchor=west] at (4.30,3.25) {$x_+$};
  \node[fflabel,text=FFGrayDark] at (3.18,0.72)
    {$\phi_s(x_-)=\phi_s(x_+)$};

  \draw[ffrule] (6.25,0.55)--(6.25,7.18);

  \node[ffneutral,minimum width=4.45cm,minimum height=0.98cm,text width=3.85cm]
    (same) at (10.25,6.35) {matched pair with the same immediate field\\$\phi_s(x_-)=\phi_s(x_+)$};
  \node[ffgold,minimum width=4.05cm,minimum height=0.92cm,text width=3.45cm]
    (grads) at (10.25,4.90) {shared future\\gradients $g_1,\ldots,g_n$};
  \draw[ffarrow] (same.south)--(grads.north);

  \coordinate (futuresplit) at (10.25,4.22);
  \draw[ffline] (grads.south)--(futuresplit);
  \node[ffblue,minimum width=2.45cm,minimum height=0.86cm]
    (minus) at (8.55,3.36) {$\phi^{(-)}_{s+n}$};
  \node[ffred,minimum width=2.45cm,minimum height=0.86cm]
    (plus) at (11.95,3.36) {$\phi^{(+)}_{s+n}$};
  \draw[ffarrowblue] (futuresplit)--(minus.north);
  \draw[ffarrowred] (futuresplit)--(plus.north);

  \coordinate (futuremerge) at (10.25,2.52);
  \draw[ffline] (minus.south)--(futuremerge);
  \draw[ffline] (plus.south)--(futuremerge);
  \node[ffred,minimum width=5.35cm,minimum height=1.10cm,text width=4.75cm]
    (derivative) at (10.25,1.32) {$\displaystyle
      \left.\partial_\xi\phi_{s+n}\right|_{0}
      =\phi_{s+n}(0)(\mu_n-\nu_n)$};
  \draw[ffarrowred] (futuremerge)--(derivative.north);

  \node[ffsubtle,text width=7.3cm] at (10.25,0.25)
    {revelation is future-gradient and context conditioned;\\cross-bank directional coherence is an additional hypothesis};
\end{tikzpicture}}
    \caption{Adam matched-section cancellation and future-gradient revelation.}
    \label{fig:adam-revelation}
\end{figure}
\FloatBarrier

\section{Global transport and resolution-supported geometry}
\label{sec:transport}

\subsection{Canonical directed transport}

The quotient functor of \cref{thm:quotient-descent} supplies canonical directed transport along every controlled protocol. For a loop $\gamma:c\to c$, the descended endomorphism belongs to a monodromy semigroup. Only on a reversible subgroupoid do loops act as group-valued holonomy.

\begin{theorem}[Endpoint independence and trivial holonomy]
\label{thm:path-independence}
On a connected context groupoid whose quotient transports are isomorphisms, the following are equivalent:
\begin{enumerate}[label=(\roman*)]
\item transport between two contexts depends only on endpoints;
\item every loop has identity holonomy;
\item after choosing a base context, the quotient functor admits a path-independent trivialization.
\end{enumerate}
\end{theorem}

\begin{proof}
Endpoint dependence makes every loop the identity. Conversely, choose one path from the base to each context; identity loop holonomy makes the induced transport independent of that choice. The trivialization immediately implies endpoint dependence.
\end{proof}

The set-level theorem does not supply a smooth connection. A vector or Hilbert bundle requires constant-rank or local-triviality hypotheses, or norm-continuous projection fields. A differential connection additionally requires smooth thin-path transport. Rank jumps yield only stratumwise geometry.

\subsection{Resolution-supported empirical transport}

Empirical linear transport is often reconstructed from a source response frame $X:E\to Y_0$ and a target frame $Z:E\to Y_1$, with $E,Y_0,Y_1$ finite-dimensional Hilbert spaces. If $X$ has unresolved small singular values, the ambient estimate $ZX^\dagger$ is unstable. The appropriate object is transport on a declared resolved source support.

For $\delta>0$ separated from the singular spectrum of $X$, define
\begin{equation}
P_\delta:=\mathbf 1_{[\delta^2,\infty)}(X^*X),
\qquad
E_\delta=P_\delta E.
\label{eq:resolved-projector}
\end{equation}
Thus $X|_{E_\delta}$ has singular values at least $\delta$.

\begin{definition}[Resolution-supported transport]
\begin{equation}
T^{(\delta)}_{Z\leftarrow X}
:=ZP_\delta(XP_\delta)^\dagger:Y_0\to Y_1.
\label{eq:supported-transport}
\end{equation}
The map acts exactly on $X(E_\delta)$ and vanishes on its orthogonal complement.
\end{definition}

\begin{theorem}[Exactness and local stability]
\label{thm:supported-transport}
The map in \cref{eq:supported-transport} is the unique linear map on $X(E_\delta)$ satisfying
\begin{equation}
T^{(\delta)}_{Z\leftarrow X}Xu=Zu
\qquad \forall u\in E_\delta.
\end{equation}
Its zero extension is the minimum-Hilbert--Schmidt-norm exact ambient extension. Moreover, if $(\widetilde X,\widetilde Z)$ is sufficiently close to $(X,Z)$ and preserves the retained dimension and spectral gap around $\delta$, then the supported transport varies locally Lipschitz-continuously: there exists a constant $C$, depending only on the resolved singular scale, the spectral gap, and a local bound on $Z$, such that
\begin{equation}
\|T^{(\delta)}_{\widetilde Z\leftarrow\widetilde X}
-T^{(\delta)}_{Z\leftarrow X}\|
\le C\bigl(\|\widetilde X-X\|+\|\widetilde Z-Z\|\bigr).
\label{eq:supported-transport-bound}
\end{equation}
\end{theorem}

\begin{proof}[Proof sketch]
On $E_\delta$, the source map is injective with pseudoinverse norm at most $\delta^{-1}$. Exactness and uniqueness on the resolved range follow from $(XP_\delta)^\dagger XP_\delta=P_\delta$. Every exact extension differs by an arbitrary map on $X(E_\delta)^\perp$, so the zero extension has minimum Hilbert--Schmidt norm. Under a fixed retained rank and singular gap, standard singular-subspace and Moore--Penrose perturbation results imply local Lipschitz continuity \citep{wedin1972perturbation,stewart1990matrix}.
\end{proof}

\begin{remark}[Conditioning scale]
Perturbation constants necessarily deteriorate as the resolved singular scale approaches zero; typical terms scale as $\delta^{-1}$ for target perturbations and $\delta^{-2}$ for source-pseudoinverse perturbations. The exact constants depend on the chosen gap theorem.
\end{remark}

\begin{theorem}[Ambient no-go at vanishing resolution]
\label{thm:transport-nogo}
There is no uniform perturbation-stable extension of $ZX^\dagger$ over source families whose smallest retained singular value tends to zero. Hence an unresolved direction cannot be interpreted as identity transport, zero monodromy, or path independence.
\end{theorem}

\begin{proof}
Take $X_t=\diag(1,t)$ and fixed $Z=I$. Then $ZX_t^\dagger=\diag(1,t^{-1})$, whose norm diverges as $t\downarrow0$ although $X_t\to X_0$.
\end{proof}

\subsection{Partial monodromy}

For a path $\gamma$ composed of empirical edges, supported transports can be composed only on the common admissible domain surviving all edge resolutions. The resulting $M_\gamma^{(\delta)}$ is a partial monodromy operator. Failure to define it on an unresolved direction is different from proving that the loop acts as the identity there.

\subsection{Context-independent propagation as an additional factorization}

A fixed action-labelled hidden propagator $Q_a$ exists only when predictive transport factors naturally through the action-label monoid. Same-label transport discrepancies or label-null loop effects obstruct that factorization. Consequently, failure of a context-independent $Q_a$ is compatible with a valid context-indexed quotient functor.

\begin{boundary}[Global geometry]
The unconditional global object is the category of elements of the predictive quotient functor. Smooth bundles, differential connections, globally fixed action propagators, and nontrivial holonomy are additional realizations or empirical memberships; none is required to define fiber fingerprints.
\end{boundary}

\section{Transformer studies and empirical membership}
\label{sec:experiments}

\subsection{Execution contexts and evidentiary discipline}

The empirical program uses frozen Qwen2.5-7B and Mistral-7B-v0.3 base models \citep{qwen2024report,jiang2023mistral}, low-rank adapters \citep{hu2022lora}, AdamW execution states \citep{kingma2015adam,loshchilov2019adamw}, three text domains, and controlled task histories. Model and data revisions, token windows, random seeds, precision contracts, probe panels, selection rules, and CPU verdict scripts were fixed before protected scientific outcomes were inspected. Accelerator notebooks generated raw artifacts only; confirmatory statistics were computed by separate frozen CPU analyses. Failed criteria were not repaired by pooling, threshold revision, post-outcome feature selection, or replacing the estimand.

The complete execution-state semantics follow \cref{thm:transformer-instantiation}. The studies test membership in conditional layers; they do not serve as premises for the finite predictive quotient.

\subsection{Overview of the evidence chain}

\Cref{tab:empirical-map} summarizes the evidence using descriptive study names. Internal archival keys are given only in \cref{app:reproducibility}. The sequence begins by identifying a local action backbone and its longer-horizon non-closure, then independently replicates visible-relative completion, and finally tests stronger transport, switching, accessibility, and revelation claims.

\begin{table}[tbp]
\centering
\caption{Empirical membership of the conditional theory.}
\label{tab:empirical-map}
\small
\begin{tabularx}{\linewidth}{@{}L{0.24\linewidth}Y L{0.17\linewidth}@{}}
\toprule
Empirical object & Evidence supported by the frozen studies & Status \\
\midrule
Local action and finite return & Action-specific one-step backbone; longer-horizon first-return non-closure. & \textbf{Supported, local} \\
Visible-relative completion & Fresh histories reproduce visible response, output-range reuse, and a low-rank irreducible-new sector in both models. & \textbf{Supported} \\
Re-anchored path transport & No path effect is resolved above the independent-bank floor; most ambient relative transports are ill conditioned. & \textbf{Not resolved} \\
Hard-switching chronology & Crossing, Hodge, bridge, and smooth-control diagnostics support the mechanism and prospective contraction. & \textbf{Strict conjunction unmet} \\
Adam accessibility & Mistral matched sections are accessible in the frozen chart; joint Qwen--Mistral panel membership is not established. & \textbf{Context/metric dependent} \\
Future-gradient revelation & Nonzero responses, zero-gradient controls, amplitude scaling, and analytic recurrence support de-cancellation. & \textbf{Supported in Mistral} \\
Cross-bank direction & Future-gradient banks can reveal the same hidden difference in different or opposite parameter-space directions. & \textbf{Bank invariance unsupported} \\
\bottomrule
\end{tabularx}
\end{table}

\subsection{Local action backbone and first-return non-closure}

In current-state-defined local frames, fresh Qwen and Mistral histories support an action-specific approximate law
\begin{equation}
C_{a,c}=Y_{a,c}X_c^\dagger\approx B_a.
\label{eq:local-action-law}
\end{equation}
The subsequent return studies show that this one-step backbone is not a closed finite-dimensional cocycle. Ordered two-step response contains a first-return term
\begin{equation}
D_{b|a,c}=C_{b|a,c}C_{a,c}+K_{b\leftarrow a,c}.
\label{eq:first-return-emp}
\end{equation}
The full $9\times9$ first-return system has local $r_{90}$ and $r_{95}$ near four over horizons H3--H5, and shifted-H6 low rank replicates. A fixed context-independent action propagator and a resolved ambient moving-fiber geometry are not supported.

The conclusion is structural rather than merely negative: a low-dimensional local response backbone exists, but longer-horizon response is history-conditioned and contains returns not exhausted by the declared current action core.

\subsection{Fresh visible-relative completion replication}

\begin{figure}[tbp]
\centering
\includegraphics[alt={Stacked horizontal bars compare visible, visible-range reuse, and irreducible-new response-energy fractions at source horizon H5 and shifted horizon H6.},width=0.98\linewidth]{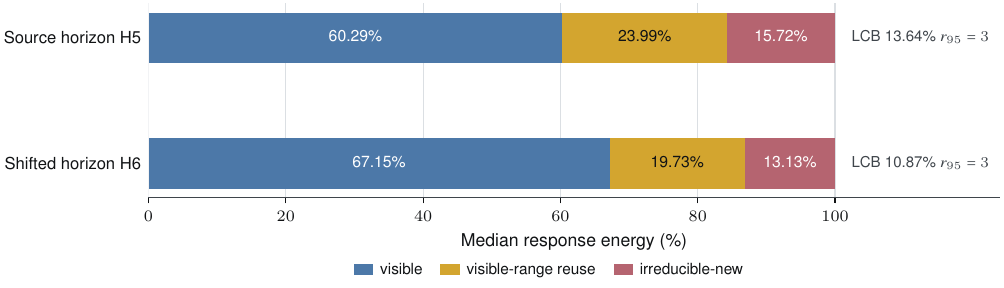}
\caption{Fresh visible-relative completion at H5 and shifted H6.}
\label{fig:completion-sectors}
\end{figure}

The completion replication uses new histories and seeds, a frozen visible projector, independent readout banks, and a predeclared eligibility-first replacement rule. The result is supported jointly in Qwen and Mistral without pooling rescue. At source horizon H5, the median response energy decomposes as
\begin{equation}
60.29\%\ \text{visible}
+23.99\%\ \text{reuse}
+15.72\%\ \text{irreducible-new},
\end{equation}
with a one-sided 95\% lower bound of $13.64\%$ on the irreducible-new fraction. At shifted H6,
\begin{equation}
67.15\%\ \text{visible}
+19.73\%\ \text{reuse}
+13.13\%\ \text{irreducible-new},
\end{equation}
with lower bound $10.87\%$. The median $r_{95}(S)$ is three at both horizons, and no source-panel unit requires more than four dominant residual modes. The frozen sector composition is shown in \cref{fig:completion-sectors}.

Thus the finite-sector picture is a declared three-dimensional visible action core, structured visible-output reuse, one or two dominant irreducible modes, and a weaker residual tail. This is a response-completion statement, not a claim about physical neural-state dimension.

\subsection{Re-anchored path-transport boundary}

The path-transport study compares AB and BA histories after exact AA+C re-anchoring, with two independent endpoint readout banks and twelve valid model-history units. The mean cross-bank energy is positive, $3.57\times10^{-6}$, but its one-sided 95\% lower bound is $-3.00\times10^{-6}$, the synchronized sign $p$-value is $0.171875$, and the median signal-to-noise ratio is $0.001651$. Only one of twelve relative target-frame reconstructions passes the principal conditioning criterion.

The preregistered nontrivial-transport claim is therefore not supported. The result does not establish path independence: the declared panel does not resolve a path-conditioned effect, and its ambient relative transport is itself unstable on most units. This distinction is exactly the one formalized by \cref{thm:supported-transport,thm:transport-nogo}.

\subsection{Controlled hard-switching scaling and finite-scale bias}

The switching studies use a frozen smooth Transformer feature map with a trainable hard-ReLU head and a softplus control. They directly record crossing counts, event marks, graph-Hodge cycle response, two future-response bridges, and smooth-control scaling.

\begin{figure}[tbp]
\centering
\includegraphics[alt={Log-scale dumbbell plot showing that absolute exponent deviations for crossing count, cycle response, two future bridges, and the softplus control contract from the initial grid to the refined grid.},width=0.98\linewidth]{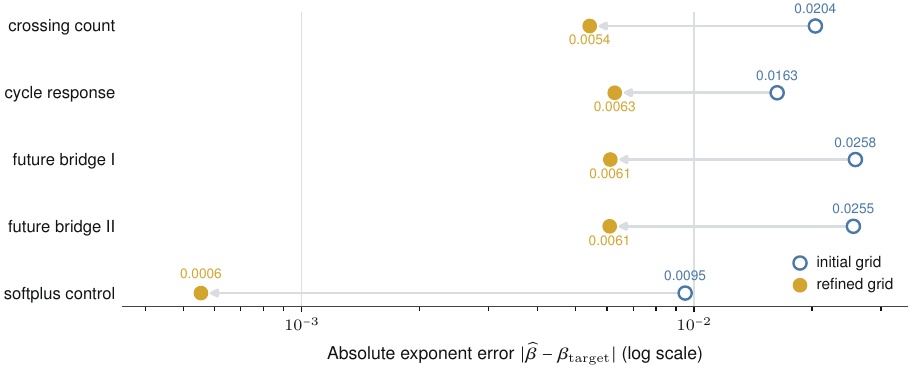}
\caption{Contraction of finite-grid exponent errors after prospective refinement.}
\label{fig:switching-convergence}
\end{figure}

On the initial grid, the cycle response scales approximately as $p^{0.5113}\eta^{1.5163}$ and the softplus control as $\eta^{2.0095}$. A prospectively refined grid uses smaller $\eta$ and larger width. Its primary cycle estimates are
\begin{equation}
\beta_p^{\mathrm{cyc}}=0.497334
\quad[0.493372,0.501296],
\qquad
\beta_\eta^{\mathrm{cyc}}=1.506284
\quad[1.503177,1.509391].
\end{equation}
Future-bridge $\eta$ exponents are approximately $1.5061$, while the softplus exponent is $2.000554$. The primary Hodge-cycle equivalence intervals are satisfied. The complete preregistered conjunction is nevertheless not met because four narrow finite-grid boundaries are exceeded by $2.8\times10^{-4}$ to $3.6\times10^{-4}$. The prospective contraction is shown in \cref{fig:switching-convergence}.

The appropriate conclusion is mixed: strong prospective support for the controlled hard-ReLU Hodge--$3/2$ mechanism and the second-order law in \cref{prop:second-order-law}, but no claim that native SiLU Transformer training belongs to the same universality class.

\subsection{Matched-section accessibility and future revelation}

The joint accessibility screen first asks whether natural history directions align with the exact Adam matched-section tangent and whether a frozen two-step, three-coordinate gradient-control chart reaches a section point in both models. Natural alignment is high in both contexts. Mistral satisfies the frozen accessibility screen for all six tested candidate seeds at the first slot; Qwen satisfies none under the declared raw $(m,v)$ Euclidean conditioning criterion. Because the joint eligibility firewall stops before selection, no protected future gradients are computed. The result is therefore a joint finite-panel accessibility failure, not a negative result about future revelation and not a coordinate-free inaccessibility theorem.

A separate prospective study then tests revelation only in the empirically accessible Mistral subclass across six frozen chronology slots. All six slots produce dynamically reconstructed matched sections. All twelve full-amplitude future-bank responses are nonzero, every full-amplitude zero-gradient control is exactly zero, and the median full/half norm ratio is $1.99895$ with median direction cosine $0.99999894$. The analytic Adam recurrence accurately reconstructs the observed response. These are direct mechanism-level signatures of future-gradient de-cancellation.

\begin{figure}[tbp]
\centering
\includegraphics[alt={Panel a compares half- and full-amplitude matched-section response norms across six Mistral chronology slots; panel b compares future-gradient-bank cosine with response cosine.},width=0.99\linewidth]{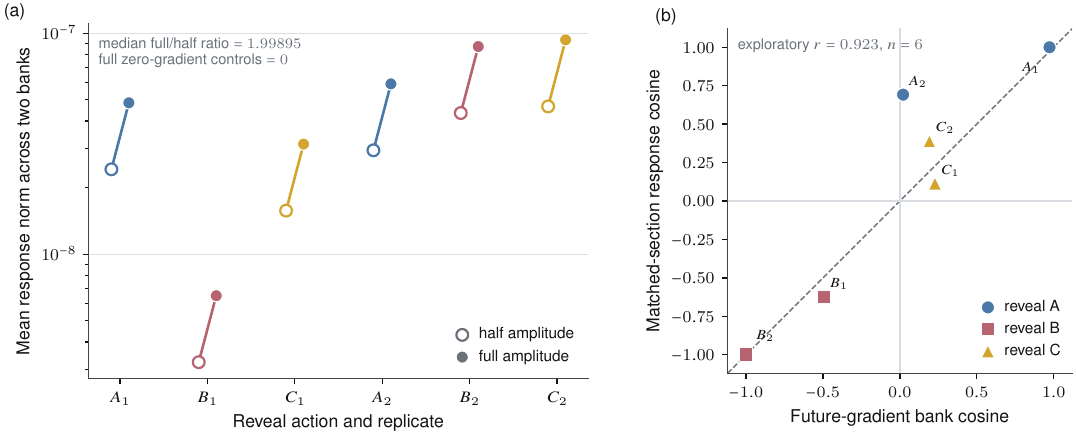}
\caption{Matched-section response scale and cross-bank direction; panel (b) is exploratory.}
\label{fig:revelation-summary}
\end{figure}

The stronger cross-bank coherence hypothesis is not supported. Four of six cross-bank inner products are positive, the one-sided lower bound is negative, and the response direction follows the geometry of the particular future-gradient bank. One of twenty-four analytic reconstruction cells also narrowly misses its 1\% relative-error criterion. The study therefore supports dynamic accessibility and future-gradient-conditioned revelation in Mistral, but not a bank-invariant revelation direction. \Cref{fig:revelation-summary} separates the preregistered amplitude evidence from the exploratory directional geometry.

\subsection{Empirical synthesis}

Together the studies support three nested statements:
\begin{enumerate}
\item present behavior, and even a declared initial learning action, can fail to determine longer-horizon response;
\item the remaining response contains visible-output reuse and a small irreducible completion on natural-history support;
\item exact optimizer-state cancellation can hide differences that controlled future gradients subsequently reveal.
\end{enumerate}
They do not establish a globally smooth empirical bundle, a universal finite-rank state, native Transformer $3/2$ membership, nontrivial path monodromy, coordinate-free Qwen inaccessibility, or bank-invariant revelation. The positive and negative results jointly identify the supported region of the theory rather than requiring post hoc repair of the finite predictive core.

\section{Related work and novelty boundary}
\label{sec:related}

The framework combines established mathematical ingredients. Its novelty claim concerns their coupling to present-behavior equivalence and controlled future learning, not the invention of those ingredients in isolation.

\subsection{Observability and predictive state}

Nonlinear observability asks whether internal states can be distinguished from future input--output behavior \citep{hermann1977nonlinear}. Predictive-state representations encode state through predictions of future action-conditioned observations \citep{littman2001predictive}. Fiber fingerprints share that operational spirit but specialize it to learning systems: the state is a complete training execution state, present behavior is explicitly quotiented, and the distinguishing experiments are future \emph{training} protocols rather than ordinary environment actions or passive outputs. The categorical quotient is defined before any smooth observability rank condition. Its minimality is related to Nerode-style behavioral quotients \citep{nerode1958linear}, while the global set-level assembly uses the standard category-of-elements construction \citep{maclane1998categories}.

The closest recent conceptual result is the distinction between loss-visible and loss-invisible overlaps in low-rank recurrent networks \citep{ger2026invisible}. Loss-invisible overlaps can encode training history and can be revealed by learning even when functionally equivalent networks initially agree. That result strongly supports the general premise that learning-relevant state can be invisible to current loss. Our framework differs in scope and construction: it is optimizer- and architecture-agnostic at the finite core, defines equivalence through arbitrary declared present readouts, treats controlled future training protocols as a probe doctrine, and adds quotient minimality, history-reachable visible-relative completion, and exact Transformer--AdamW instantiation.

\subsection{Function-preserving gauges and hidden learning dynamics}

Function-preserving parameter transformations need not preserve training dynamics. Path-conditioned rescaling exploits positive-homogeneous ReLU symmetries to change optimization conditioning while keeping the represented function fixed \citep{lebeurrier2026path}. Hidden-gauge analysis further shows that a predictor-preserving scaling choice can determine which unit acquires a feature and can produce large separations in specialization time \citep{wang2026hiddengauge}. These studies demonstrate that same-function parameterizations need not have the same future trajectory. Fiber Theory does not claim that broad fact as new. Its distinct target is a protocol-relative predictive quotient of complete learning execution states, including cases in which parameters and immediate adaptive fields are also matched.

\subsection{Balancing, shorting, and predictive completion}

Past-generation and future-observation energy are classical themes in dissipativity and nonlinear balancing \citep{willems1972dissipative,scherpen1993balancing}; empirical balancing also constructs reduced models from finite perturbation data \citep{lall2002subspace}. Moore--Penrose inversion, Douglas range inclusion, shorted operators, and optimal low-rank approximation are established tools \citep{penrose1955generalized,douglas1966majorization,anderson1975shorted,eckart1936approximation}. Our use of these tools is relative to a frozen visible learning-action backbone: input-invisible response is separated into visible-output reuse and irreducible-new directions, and the factorization obstruction is then weighted by the covariance of naturally reachable histories. The result is a response-completion dimension, not a state-space identification theorem.

\subsection{Task order, Lie brackets, and optimizer memory}

Graph Hodge decomposition of edge flows is standard \citep{jiang2011hodge}. Smooth local order effects are governed by Lie brackets, and recent work uses such brackets to analyze multi-domain order and predict transfer order \citep{rukhovich2025commute,sweeney2026geometry}. Trained Transformer adaptation can be first-order predictable along individual directions while pairwise composition and task order remain fragile \citep{piontkovskaia2026fragile}. Optimizer memory can also elevate equal-multiset shuffle-order contrasts from second to first order in the learning rate \citep{sweeney2026memory}. These results are closely related to our empirical local-backbone/non-closure picture.

The distinctive chronology claim here is narrower and mechanistic: under a common-affine memory channel, graph-cycle projection exactly removes task-potential memory and isolates state-dependent interaction. Under a separately declared crossing regime, that cycle field inherits a graph-projected $3/2$ rare-event law with a cyclomatic prefactor. Neither the Hodge decomposition nor the smooth Lie bracket is claimed as new.

\subsection{Adam, adapters, and execution-state geometry}

Adam and AdamW recurrences are standard \citep{kingma2015adam,loshchilov2019adamw}; LoRA and quantized LoRA provide the parameter-efficient execution contexts used in the experiments \citep{hu2022lora,dettmers2023qlora}. Our optimizer contribution is an exact moment-space section preserving the immediate Adam adaptive field, an exact scalar moment-reachability region, a separation of neural and metric-stable accessibility, and a closed first-order formula for future-gradient de-cancellation. This is an execution-state fiber even when the trainable parameter vector is held fixed.

\subsection{Plasticity and future-learning probes}

Recent work operationalizes plasticity through future optimization gain and studies diagnostics for future trainability \citep{wang2026plasticity}. Short domain-specific probing has also been used to estimate cross-domain effects for constrained LLM fine-tuning \citep{gualdoni2026dynamics}, and few-shot adaptation has been proposed as a richer stability/plasticity evaluation than zero-shot recall alone \citep{inamdar2026reevaluating}. These works support the practical relevance of future-learning probes. The present paper focuses on the state-theoretic question that precedes a deployment decision: which distinctions exist inside present-behavior equivalence classes, how are they generated, and through which future-learning operators can they be revealed?

\subsection{Models and experimental context}

The empirical contexts instantiate the Transformer architecture \citep{vaswani2017attention} through the Qwen2.5 and Mistral model families \citep{qwen2024report,jiang2023mistral}. These citations identify the base architectures; the claims in \cref{sec:experiments} are specific to the frozen model revisions, LoRA instrumentation, task domains, precision contracts, and probe panels recorded in the artifact release.

\subsection{Precise novelty statement}

The paper does not claim to invent observability, predictive-state representations, graph Hodge decomposition, Lie brackets, pseudoinverses, shorted operators, balancing, dissipativity, or Adam. Its contribution is the coupled architecture
\begin{equation}
\begin{aligned}
&\text{present-behavior-matched learning execution states}
+\text{future training as the probe}\\
&+\text{a minimal predictive quotient and set-level global fiber}
+\text{history-generated occupancy}\\
&+\text{visible/reuse/irreducible-new completion}
+\text{chronology and optimizer cancellation/revelation}.
\end{aligned}
\label{eq:novelty-architecture}
\end{equation}
The strongest theorem-level contributions within this architecture are the controlled-learning predictive quotient and its minimality, the history-reachability--visible-completion bridge, the common-affine coboundary--interaction decomposition, the graph-projected switching law, and the exact Adam matched-section cancellation and future-gradient derivative. The defensible novelty claim is that no direct prior formulation of this full coupled architecture was located in our review, not an unqualified worldwide-first claim.

\section{Limitations, non-implications, and open boundaries}
\label{sec:limits}

The theory is intentionally modular. This section states what is not licensed by either the mathematics or the experiments.

\subsection{Probe relativity and measurement}

Every predictive quotient and fingerprint is relative to a declared probe family, response gauge, horizon, and measurement semantics. A finite bank need not be complete. Approximate equality within a tolerance is generally not transitive and is not an exact quotient unless a congruent partition or stability theorem is declared. Population response, finite readout, and implemented floating-point response are distinct objects.

\subsection{Regularity and representation}

The set-level fiber need not be a manifold. Tangent spaces, Gramians, pseudoinverses, spectra, and shorted operators require local regularity, stable rank or closed range, and declared metrics. The main completion theorems are stated in finite dimension. Infinite-dimensional analogues require trace-class history covariance, closed-range or Douglas factorization hypotheses, and care with unbounded pseudoinverses. Low response rank cannot identify physical neural-state dimension.

A smooth global bundle or connection requires local triviality and smooth path transport; local low rank is insufficient. The unconditional global object remains the category of elements of the predictive quotient functor.

\subsection{Transport}

The re-anchored path study rejects one specific empirical transport membership. It does not prove global path independence. Conversely, the abstract quotient functor does not guarantee that a finite empirical response frame identifies a stable ambient linear transport. Resolution-supported partial transport is the appropriate finite estimand when source directions approach the measurement floor.

\subsection{Switching and the \texorpdfstring{$3/2$}{3/2} class}

The $3/2$ exponent is conditional on crossing count, mark scale, centering, finite variance, aggregate dependence, and the joint limit $p\eta\to\infty$. The controlled studies use a hard-ReLU head over frozen Transformer features; they do not establish native Qwen or Mistral SiLU membership. The preregistered finite-grid conjunction was not met. The second-order proposition explains how finite-grid slopes may approach the same asymptotic class, but it does not retroactively alter the empirical verdict.

\subsection{Optimizer reachability and revelation}

An algebraic matched section is not necessarily visited by natural neural training. Optimizer-level moment reachability is not neural-data reachability. Raw-coordinate condition numbers are metric dependent. The joint accessibility screen does not prove coordinate-free Qwen inaccessibility. The conditional Mistral study supports dynamic accessibility and future-gradient-conditioned revelation but does not establish a stable bank-invariant response direction. The closed derivative formula describes the first resolving event under exogenous common gradients; longer endogenous dynamics require the full coupled tangent recursion.

\subsection{Generalization}

The empirical evidence is limited to the declared Qwen and Mistral revisions, LoRA parameterization, AdamW conventions, text domains, horizons, and numerical contracts. Cross-model, cross-optimizer, or cross-backend claims require independent replication. A CUDA implementation and a TPU/XLA implementation are distinct execution contexts unless an exact or tolerance-controlled intertwiner is established.

\subsection{Open questions}

Open mathematical and empirical questions include complete finite-probe criteria in nonlinear or infinite-rank systems, all-horizon rank stabilization, native Transformer switching membership, supported cross-context transport, and independent replication on additional model and optimizer families. These are extensions of the present theory rather than missing premises of the finite predictive quotient.

\subsection{Audit status}

The theorem chain, counterexamples, source code, hash manifests, and frozen analyses have undergone internal adversarial review and repeated computational checks. They have not undergone external peer review or proof-assistant certification. The release therefore provides a formal manuscript and reproducibility archive, not a claim of mechanized proof completeness.

\begin{table}[H]
\centering
\caption{Key non-implications retained by the theory.}
\small
\begin{tabularx}{\linewidth}{@{}Y c Y@{}}
\toprule
Premise & & Not implied \\
\midrule
Present-behavior equality & $\not\Rightarrow$ & future-learning equality \\
Set-level quotient transport & $\not\Rightarrow$ & stable ambient linear transport \\
Asymptotic $3/2$ membership & $\not\Rightarrow$ & exact finite-grid exponent \\
Algebraic moment reachability & $\not\Rightarrow$ & neural-data accessibility \\
Surjective control differential & $\not\Rightarrow$ & arbitrary raw-coordinate condition threshold \\
Positive future observability & $\not\Rightarrow$ & positive cross-bank directional coherence \\
Low completion rank & $\not\Rightarrow$ & low physical neural-state dimension \\
\bottomrule
\end{tabularx}
\end{table}

\section{Conclusion}
\label{sec:conclusion}

A learning state is not determined solely by its present behavior. Relative to a declared family of future-training probes, present-behavior equivalence classes carry structured future-response laws. The finite controlled experiment gives these laws a canonical predictive quotient, a minimal recursively sufficient representation, and a global set-level fiber without assuming smoothness or fixed dimension.

Under explicit realizations, this core supports a richer but still modular structure. Visible-relative completion separates declared action response, visible-output reuse, and irreducible-new modes. The history-reachability bridge restricts that geometry to directions actually generated and occupied by training histories. Graph-Hodge chronology distinguishes task-potential memory from state-dependent interaction, while a conditional crossing regime yields a graph-projected $3/2$ scale with finite-scale corrections. Adam supplies an exact execution-state fiber: different moment states can share the same immediate adaptive field, and common future gradients can locally reveal their hidden difference.

The Transformer studies support this hierarchy while also locating its boundaries. Local action response is structured but not closed at longer horizons; fresh visible-relative completion contains both reuse and a low-rank irreducible sector; the declared re-anchored transport panel is unresolved; the controlled switching mechanism approaches but does not satisfy its full finite-grid conjunction; and accessible Mistral matched sections are future-gradient revealing but not bank-invariant. These outcomes motivate a geometry that is support-conditioned, scale-resolved, metric-resolved, and context-indexed.

The principal conclusion is therefore precise:
\begin{equation}
\text{present behavior is not a sufficient statistic for declared future learning.}
\end{equation}
Fiber fingerprints provide an operational way to represent the missing predictive structure. Determining which parts of that structure should be acquired and used for a particular production decision is a downstream decision-theoretic problem, not a prerequisite for the state theory developed here.

\appendix
\section{Proof details for the finite predictive core}
\label{app:proofs-core}

This appendix records the complete elementary arguments behind the core theorem chain. The core uses only sets, functions, finite protocols, and prefix-compatible response semantics.

\subsection{Proof of predictive descent and functoriality}

Let $q_c:X_c\to Q_c^\infty$ be the quotient map. For an arrow $a:c\to d$, define
\begin{equation}
\tau_a(q_c(x)):=q_d(T_ax).
\end{equation}
If $q_c(x)=q_c(y)$, then $x\sim_c^\infty y$. For each $q\in\cP(d)$,
\[
\Sigma_{d,q}(T_ax)=\Sigma_{c,a^*q}(x)=\Sigma_{c,a^*q}(y)=\Sigma_{d,q}(T_ay),
\]
so $q_d(T_ax)=q_d(T_ay)$. Hence $\tau_a$ is well defined and unique because $q_c$ is surjective. Moreover,
\begin{align}
\tau_{\mathrm{id}_c}(q_c(x))&=q_c(T_{\mathrm{id}_c}x)=q_c(x),\\
\tau_b\tau_a(q_c(x))&=q_e(T_bT_ax)=q_e(T_{b\circ a}x)=\tau_{b\circ a}(q_c(x)).
\end{align}
Thus $Q^\infty$ is a functor.

\subsection{Grothendieck construction}

The category $\int_{\cC}Q^\infty$ has objects $(c,z)$ with $z\in Q_c^\infty$. A morphism over $a:c\to d$ from $(c,z)$ is uniquely
\begin{equation}
(c,z)\longrightarrow(d,\tau_a z).
\end{equation}
This is the opcartesian lift required for a discrete opfibration. Composition and identities follow from functoriality. No topology or rank assumption is involved.

\subsection{Nerode minimality}

Suppose every response factors as $\Sigma_{c,q}=\bar\Sigma_{c,q}\circ z_c$. If $z_c(x)=z_c(y)$, then all response values agree, hence $q_c(x)=q_c(y)$. Define
\begin{equation}
\bar q_c:z_c(X_c)\to Q_c^\infty,
\qquad
\bar q_c(z_c(x))=q_c(x).
\end{equation}
The implication just proved makes $\bar q_c$ well defined. It is unique because $z_c:X_c\to z_c(X_c)$ is surjective. If controlled transitions descend on $Z_c$, the maps $\bar q_c$ intertwine them with $\tau_a$, since both compositions agree on every $z_c(x)$.

\subsection{Finite-horizon loss of depth}

Let $q$ have horizon at most $H$ at $d$, and let $a:c\to d$ consume $\ell$ steps. Then $a^*q$ has horizon at most $H+\ell$. Therefore $x\sim_c^{\le H+\ell}y$ implies
\begin{equation}
\Sigma_{d,q}(T_ax)=\Sigma_{c,a^*q}(x)=\Sigma_{c,a^*q}(y)=\Sigma_{d,q}(T_ay).
\end{equation}
This proves the map $Q_c^{\le H+\ell}\to Q_d^{\le H}$ and explains why equal horizon labels on both sides are generally mistyped.

\subsection{Probe enrichment}

If $A\subseteq A'\subseteq\cP(c)$ and two states agree on all probes in $A'$, they agree on $A$. Hence the $A'$ quotient refines the $A$ quotient. The same argument proves that cumulative horizon quotients refine monotonically as $H$ grows.

\subsection{Depth filtration}

For a fixed compatible control prefix, the map $C_\alpha^{(k+1)}$ contains every component of $C_\alpha^{(k)}$. Hence equality under $C_\alpha^{(k+1)}$ implies equality under $C_\alpha^{(k)}$ and
\begin{equation}
F_{x,\alpha}^{(k+1)}\subseteq F_{x,\alpha}^{(k)}.
\end{equation}
No differential argument is needed.

\subsection{Regular tangent specialization}

If $C_\alpha^{(k)}$ is $C^1$ and has locally constant rank near $x$, the constant-rank theorem \citep{lee2013smooth} gives coordinates in which it is a projection. The local level set is a submanifold and its tangent is $\Ker D(C_\alpha^{(k)})_x$. At a singular point only the curve tangent cone is guaranteed to lie in this kernel. For example, the closed set $[0,\infty)$ at zero has one-sided tangent cone $[0,\infty)$ but no nonzero reversible tangent direction. This is why centered finite probes require an admissible two-branch realization rather than an arbitrary vector in a differential kernel.

\subsection{Finite operator data are not a complete ontology}

Let $F_1(z)=Lz$ and $F_2(z)=Lz+q(z)$ with $q(0)=Dq(0)=0$ but $q\not\equiv0$. Their first derivatives and all first-order operator objects agree at zero, while finite responses differ. Therefore a first-order Hilbert realization cannot reconstruct the unrestricted finite-scale geometry without a complete response profile, analytic/jet integrability, or another explicit closure condition.

\subsection{Non-identifiability of implementation dimension}

Let $(X,T,\Sigma)$ be a controlled experiment and $(A,S)$ any auxiliary controlled system. Define the product state $(x,a)$ with transition $(T_u x,S_u a)$ and responses $\widetilde\Sigma_q(x,a)=\Sigma_q(x)$. Projection to $X$ is response sufficient, so the predictive quotient is unchanged. The auxiliary dimension and complexity are arbitrary. Operational response data can therefore identify a minimal predictive quotient but not the full implementation dimension.

\section{Proof details for completion, chronology, and optimizer mechanisms}
\label{app:proofs-mechanisms}

\subsection{Visible-relative decomposition}

Let $A_0=KP$ and $B_0=KQ$. The visible output projector is $\Pi=A_0A_0^\dagger$. Then
\begin{equation}
K=A_0+\Pi B_0+(I-\Pi)B_0.
\end{equation}
The first two input blocks are orthogonal because $PQ=0$. The last block is output-orthogonal to the first two because $(I-\Pi)\Pi=0$ and $(I-\Pi)A_0=0$. Hence the decomposition is Hilbert--Schmidt orthogonal.

Write
\begin{equation}
A=A_0^*A_0,
\qquad
C=A_0^*B_0,
\qquad
D=B_0^*B_0.
\end{equation}
Using $A_0(A_0^*A_0)^\dagger A_0^*=\Pi$ on the closed finite-dimensional range,
\begin{equation}
C^*A^\dagger C=B_0^*\Pi B_0,
\end{equation}
so
\begin{equation}
D-C^*A^\dagger C=B_0^*(I-\Pi)B_0=K_{\mathrm{new}}^*K_{\mathrm{new}}.
\end{equation}
This proves the shorted identity.

For any $T$, $KPT$ lies in $\Ran(KP)$. The orthogonal projection theorem therefore gives
\begin{equation}
\|KQ-KPT\|_{\HS}^2
\ge\|(I-\Pi)KQ\|_{\HS}^2,
\end{equation}
with equality at $KPT=\Pi KQ$. The optimal rank-$m$ completion then follows from Eckart--Young applied to $(I-\Pi)KQ$.

\subsection{History bridge}

Apply the preceding result to $B_0=KQJR^{1/2}$. The visible approximants are $KPLR^{1/2}$, and the orthogonal residual is
\begin{equation}
(I-\Pi)KQJR^{1/2}=NJR^{1/2}.
\end{equation}
Thus the minimal squared error is
\begin{equation}
\|NJR^{1/2}\|_{\HS}^2
=
\tr[(NJR^{1/2})^*(NJR^{1/2})]
=
\tr\Birr.
\end{equation}
The spectral tail statement again follows from Eckart--Young.

\subsection{Monotone cumulative shorting}

For each $u$,
\begin{equation}
\langle u,S_{\le H}u\rangle
=
\inf_{v\in P\cU}\sum_{h=1}^H\|K_hQu-K_hPv\|^2.
\end{equation}
Adding the $(H+1)$st nonnegative term before taking the infimum can only increase the value. This proves $S_{\le H}\preceq S_{\le H+1}$. Congruence by $JR^{1/2}$ preserves the PSD order.

\subsection{Common-affine chronology}

Starting from
\begin{equation}
z(\Phi_a x)=Az(x)+U_a(x),
\end{equation}
write
\begin{align}
z(\Phi_b\Phi_a x)
&=A^2z(x)+AU_a(x)+U_b(x)+\Delta_aU_b(x),\\
z(\Phi_a\Phi_b x)
&=A^2z(x)+AU_b(x)+U_a(x)+\Delta_bU_a(x).
\end{align}
Their half difference is
\begin{equation}
\frac12(I-A)(U_b-U_a)
+
\frac12(\Delta_aU_b-\Delta_bU_a).
\end{equation}
The first term is the graph coboundary of $\psi_a=(I-A)U_a/2$. Hodge cycle projection annihilates it.

\subsection{Graph-projected rare-event law}

Let $P=P_{\mathrm{cyc}}\otimes I$. Then
\begin{align}
\E\|PC\|^2
&=\E\langle C,PC\rangle\\
&=\sum_{e,f}(P_{\mathrm{cyc}})_{ef}\E\langle C_e,C_f\rangle.
\end{align}
Under \cref{ass:regular-crossing}, the diagonal terms contribute
\begin{equation}
p\eta^3\sum_e(P_{\mathrm{cyc}})_{ee}q_e,
\end{equation}
and the total off-diagonal plus remainder is $o(p\eta^3)$. If $q_e=q$, the sum is $q\tr P_{\mathrm{cyc}}=qb_1(G)$.

The scaling interpretation is: $N\asymp p\eta$ centered crossing events, each with post-update displacement $O(\eta)$. Incoherent aggregation therefore has norm $O(\sqrt N\,\eta)=O(\sqrt p\,\eta^{3/2})$.

\subsection{Exact Adam moment epigraph}

Let $a=(a_1,\ldots,a_K)^T$ and $W=\diag(b_1,\ldots,b_K)$. Then
\begin{equation}
M_K=a^Tg,
\qquad
V_K=g^TWg.
\end{equation}
Weighted Cauchy--Schwarz gives
\begin{equation}
M_K^2
=\bigl\langle W^{-1/2}a,W^{1/2}g\bigr\rangle^2
\le(a^TW^{-1}a)(g^TWg)=C_KV_K.
\end{equation}
The minimum-energy solution for prescribed $M_K$ is
\begin{equation}
g_*=\frac{M_K}{C_K}W^{-1}a,
\end{equation}
with energy $M_K^2/C_K$. For $K\ge2$, choose any nonzero $h$ with $a^Th=0$ and orthogonalize it to $g_*$ in the $W$ metric. Then $g=g_*+th$ leaves $M_K$ fixed and increases $V_K$ continuously by $t^2h^TWh$, reaching every point above the boundary.

\subsection{Matched-section cancellation and future derivative}

For \cref{eq:matched-section},
\begin{equation}
\sqrt{v_s(\xi)/c_{2,s}}+\epsilon=e^\xi D_s,
\end{equation}
so
\begin{equation}
\phi_s(\xi)=\frac{e^\xi m_s/c_{1,s}}{e^\xi D_s}=\phi_s(0).
\end{equation}

Under common future gradients, the future first moment is $e^\xi M_n+G_n^{(m)}$. Its logarithmic derivative at zero is $\mu_n$. The future second moment is
\begin{equation}
A_n(e^\xi D_s-\epsilon)^2+G_n^{(v)}.
\end{equation}
Differentiation at zero gives $2A_n(D_s-\epsilon)D_s$. Applying the chain rule to the bias-corrected square-root denominator yields $\nu_n$. The logarithmic derivative of the ratio is $\mu_n-\nu_n$, proving \cref{eq:revelation-derivative}.

\subsection{Endogenous feedback boundary}

The preceding derivative treats future gradients as exogenous and common. Once the paired parameter paths diverge, gradients differ and the exact pair recursions acquire
\begin{align}
\delta m_{t+1}&=\beta_1\delta m_t+(1-\beta_1)\delta g_t,\\
\delta v_{t+1}&=\beta_2\delta v_t+(1-\beta_2)(g_t^++g_t^-)\delta g_t,\\
\delta\theta_{t+1}&=(1-\eta_t\lambda)\delta\theta_t-\eta_t\delta\phi_{t+1}.
\end{align}
Thus exogenous revelation is the first resolving event; longer endogenous dynamics require the full tangent recursion. The conditional revelation study freezes the first future-gradient comparison precisely to test the closed mechanism before endogenous feedback dominates.

\section{Resolution refinements and separation results}
\label{app:refinements}

This appendix records four refinements needed to interpret the empirical boundaries without changing the finite predictive core.

\subsection{Resolution-supported transport}

Let $X:E\to Y_0$ have singular-value decomposition $X=U\Sigma V^*$ and let $\delta$ lie in a spectral gap. Then
\begin{equation}
P_\delta=\mathbf 1_{[\delta^2,\infty)}(X^*X)
=V\mathbf 1_{[\delta,\infty)}(\Sigma)V^*.
\end{equation}
On $E_\delta=P_\delta E$, the restricted source map has pseudoinverse norm at most $\delta^{-1}$. If $\widetilde X=X+E_X$ and $\widetilde Z=Z+E_Z$ preserve the retained dimension and gap, Wedin-type subspace perturbation and Moore--Penrose perturbation give local bounds of the form
\begin{equation}
\|\widetilde P_\delta-P_\delta\|
\le c_{\mathrm{gap}}\|E_X\|,
\end{equation}
and
\begin{equation}
\|\widetilde X_\delta^\dagger-X_\delta^\dagger\|
\le c\delta^{-2}\|E_X\|+o(\|E_X\|).
\end{equation}
Consequently,
\begin{equation}
\|\widetilde T^{(\delta)}-T^{(\delta)}\|
\lesssim
\delta^{-1}\|E_Z\|
+(\|Z\|\delta^{-2}+c_{\mathrm{gap}})\|E_X\|.
\end{equation}
The exact constants depend on the selected perturbation theorem; the unavoidable feature is inverse dependence on the resolved singular scale. The family $X_t=\diag(1,t)$ proves the ambient no-go as $t\downarrow0$.

\subsection{Second-order switching law}

Suppose the graph-projected cycle energy and relative concentration satisfy \cref{ass:second-order-concentration}. Taking square roots gives \cref{eq:second-order-law}, and differentiating the logarithm gives \cref{eq:effective-exponent}. For a finite secant regression over $\eta_1<\eta_2$,
\begin{equation}
\widehat\beta_{\eta_1,\eta_2}
=\frac32+
\frac{\log a(\eta_2)-\log a(\eta_1)}{\log\eta_2-\log\eta_1},
\end{equation}
where
\begin{equation}
a(\eta)=\sqrt{\kappa_0+\kappa_1\eta+\kappa_2\eta^2+o(\eta^2)}.
\end{equation}
Thus a finite-grid exponent can sit systematically above or below $3/2$ while converging to the same asymptotic class.

\subsection{Metric-aware accessibility}

Let $J:U\to M$ and metrics $G_U,G_M\succ0$. The generalized singular values are the ordinary singular values of
\begin{equation}
\widetilde J=G_M^{1/2}JG_U^{-1/2}.
\end{equation}
Under coordinate changes $u'=Au$, $m'=Bm$ and metric transport
\begin{equation}
G'_U=A^{-*}G_UA^{-1},
\qquad
G'_M=B^{-*}G_MB^{-1},
\end{equation}
the whitened operators differ only by left and right unitary factors, so their singular values are invariant.

For \cref{prop:stable-radius}, let $J_0=DF(0)$ and let $S=(\Ker J_0)^{\perp_{G_U}}$. The metric Moore--Penrose right inverse $R_0:M\to S$ satisfies $J_0R_0=I_M$ and $\|R_0\|\le\tau^{-1}$. For $y$ near $F(0)$, define
\begin{equation}
\mathcal T_y(u)=u-R_0(F(u)-y),
\qquad u\in S.
\end{equation}
On $S$, $R_0J_0=I_S$. If $r\le\tau/(2L)$, the derivative remainder gives a contraction factor at most $Lr/\tau\le1/2$. Also $\|\mathcal T_y(0)\|\le r/2$ when $\|y-F(0)\|\le\tau r/2$. Hence $\mathcal T_y$ maps the radius-$r$ ball into itself and has a fixed point. Since $R_0$ is injective, the fixed-point equation implies $F(u)=y$, proving \cref{eq:stable-radius}.

A raw $(m,v)$ condition number can change under the coordinate scaling $v'=\lambda v$ even though rank is unchanged. A finite raw-chart conditioning result is therefore a declared metric membership, not a coordinate-free accessibility theorem.

\subsection{Revelation--coherence separation}

Let $W=\sum_b\pi_bO_b^*O_b$. Then
\begin{equation}
\Ker W=\bigcap_b\Ker O_b,
\end{equation}
because $\langle u,Wu\rangle=\sum_b\pi_b\|O_bu\|^2$. Thus $W$ is the correct positive-semidefinite object for collective visibility. Pairwise coherence instead depends on $O_b^*O_{b'}$, which need not be positive semidefinite or share a sign. The rotation example in \cref{prop:revelation-coherence} shows that complete visibility under every bank is compatible with opposite response directions.

\subsection{Four-resolution separation}

\begin{theorem}[Four-resolution separation]
\label{thm:four-resolution}
None of the following implications holds without its named additional assumptions:
\begin{align}
\text{set-level predictive transport}
&\Rightarrow\text{stable ambient linear transport},\\
\text{asymptotic $3/2$ law}
&\Rightarrow\text{exact finite-grid exponent},\\
DF\text{ surjective}
&\Rightarrow\text{a fixed raw-coordinate condition threshold},\\
\langle u,Wu\rangle>0
&\Rightarrow\langle O_bu,O_{b'}u\rangle>0.
\end{align}
\end{theorem}

\begin{proof}
The first implication is refuted by $X_t=\diag(1,t)$. The second is refuted by any nonzero $\kappa_1$ in \cref{eq:second-order-law}. The third is refuted by invertibly rescaling an output coordinate while transporting no metric. The fourth is refuted by the rotation example.
\end{proof}

The refined coupling is therefore
\begin{equation}
\text{history-generated reachable geometry}
\quad\times\quad
\text{support-, scale-, metric-, and context-resolved observability}.
\end{equation}

\section{Reproducibility contracts and study registry}
\label{app:reproducibility}

\subsection{General contract}

The empirical artifacts implement the following common discipline.
\begin{enumerate}
\item \textbf{Complete execution state.} Model parameters, LoRA state, Adam moments, optimizer step, schedules, random streams, data cursor, precision contract, and mutable runtime state are either serialized or reconstructed from pinned assets and deterministic controls.
\item \textbf{Frozen selection.} Eligibility, replacement, and candidate order are declared before protected future outcomes are computed. Rejected candidates do not generate protected outcomes.
\item \textbf{Separated analysis.} Accelerator notebooks generate raw scientific records. A separately hashed CPU script computes the frozen statistical verdict.
\item \textbf{Independent readout when required.} Transport, completion, and revelation studies use independent response banks when the estimand requires cross-bank validation.
\item \textbf{Fail-closed behavior.} Missing files, hash mismatches, invalid exact-match checks, incomplete panels, or engineering failures produce no scientific verdict.
\end{enumerate}

\subsection{Models, adapters, and domains}

The main Transformer contexts use Qwen2.5-7B and Mistral-7B-v0.3 with frozen base weights and trainable LoRA modules. AdamW execution state includes first and second moments and step counters. Three text domains, denoted A, B, and C, are instantiated by frozen datasets and disjoint training/readout token windows. Exact model, tokenizer, data, source, and analysis hashes are contained in the released experiment packages.

\subsection{Formal study registry}

The paper uses descriptive study names rather than internal development identifiers. A separate release crosswalk maps these names to archived notebooks, logs, raw packages, and frozen analysis scripts; the crosswalk is not part of the scientific naming scheme.

\small
\begin{longtable}{@{}L{0.35\linewidth}L{0.57\linewidth}@{}}
\caption{Formal study names and roles in the paper.}\label{tab:study-registry}\\
\toprule
Formal study name & Role in the paper \\
\midrule
\endfirsthead
\toprule
Formal study name & Role in the paper \\
\midrule
\endhead
Local action and return structure & Establishes an action-specific local backbone, first-return non-closure, low local return rank, and the boundary of a fixed context-independent propagator. \\
Fresh visible-relative completion replication & Independently replicates visible/reuse/irreducible-new response completion in both models. \\
Re-anchored path-transport study & Tests a finite linear transport/monodromy membership after exact endpoint re-anchoring. \\
Controlled hard-switching scaling study & Tests crossing counts, marks, graph-Hodge cycle scaling, future bridges, smooth controls, and prospective finite-scale convergence. \\
Joint matched-section accessibility screen & Tests natural tangent alignment and joint Qwen--Mistral membership in a frozen two-step, three-coordinate control chart. \\
Conditional matched-section revelation study & Tests future-gradient de-cancellation in the empirically accessible Mistral subclass. \\
\bottomrule
\end{longtable}
\normalsize

\subsection{Detailed empirical ledger}

\small
\begin{longtable}{@{}L{0.28\linewidth}L{0.52\linewidth}L{0.13\linewidth}@{}}
\caption{Frozen outcomes stated in formal language.}\label{tab:full-ledger}\\
\toprule
Study & Frozen outcome & Status \\
\midrule
\endfirsthead
\toprule
Study & Frozen outcome & Status \\
\midrule
\endhead
Local action and return structure & Fresh Qwen/Mistral histories support action-specific local maps. The one-step law extends, but a fixed closed three-dimensional two-step cocycle is not supported. The full $9\times9$ first-return system has $r_{90},r_{95}\approx4$; shifted-H6 low rank replicates. & Supported / boundary \\
Fresh visible-relative completion replication & At H5, median response energy is 60.29\% visible, 23.99\% reuse, and 15.72\% irreducible-new. At shifted H6 it is 67.15\%, 19.73\%, and 13.13\%. Both models satisfy the frozen replication without pooling rescue. & Supported \\
Re-anchored path-transport study & Mean cross-bank energy $3.57\times10^{-6}$; one-sided lower bound $-2.996\times10^{-6}$; synchronized sign $p=0.171875$; median SNR $0.001651$; one of twelve principal relative-transport condition checks passes. & Not supported \\
Controlled hard-switching scaling study & Initial cycle approximately $p^{0.5113}\eta^{1.5163}$ and softplus $\eta^{2.0095}$. Refined cycle $p^{0.497334}\eta^{1.506284}$, future bridges near $\eta^{1.5061}$, and softplus $\eta^{2.000554}$. Primary cycle equivalence holds, but four narrow conjunction criteria remain outside their frozen bands. & Mixed \\
Joint matched-section accessibility screen & Natural tangent alignment is high. Mistral satisfies the frozen first-slot screen for all six candidate seeds; Qwen satisfies none under the declared raw $(m,v)$ Euclidean conditioning criterion. No future outcome is computed. & Criterion not met \\
Conditional matched-section revelation study & Six Mistral chronology slots are dynamically reconstructed; twelve of twelve full-amplitude responses are nonzero; full zero-gradient controls are zero; median full/half norm ratio $1.99894677$; median full/half cosine $0.999998942$. A bank-invariant direction is not supported. & Mixed \\
\bottomrule
\end{longtable}
\normalsize

\subsection{Selected numerical details}

\begin{table}[H]
\centering
\caption{Median sector fractions in the fresh completion replication.}
\begin{tabular}{@{}lrrrr@{}}
\toprule
Horizon & Visible & Reuse & Irreducible-new & Median $r_{95}(S)$ \\
\midrule
Source H5 & 60.29\% & 23.99\% & 15.72\% & 3 \\
Shifted H6 & 67.15\% & 19.73\% & 13.13\% & 3 \\
\bottomrule
\end{tabular}
\end{table}
The one-sided 95\% lower bounds on the irreducible-new fraction are 13.64\% and 10.87\%, respectively. Independent projection families and action-specific separation checks address the possibility of a single-sketch artifact.

The refined hard-switching study completes sixteen frozen seeds, five widths, and six $\eta$ levels. Its four unmet conjunction criteria are: crossing-count $\eta$ upper interval endpoint $1.010357>1.010000$; two future-bridge upper endpoints $1.510310$ and $1.510316$ above $1.510000$; and a cycle point estimate $1.506284$ above a prospective band ending at $1.506000$. Validity, Hodge, centering, finite-variance, bridge, smooth-control, amplitude-prediction, and seed-stability checks are otherwise satisfied.

In the conditional revelation study, natural tangent alignment has minimum $0.919397$ and median $0.981713$. Full-amplitude future-response norms lie between $5.74\times10^{-9}$ and $1.41\times10^{-7}$. One of twenty-four analytic reconstruction comparisons narrowly misses a 1\% relative-error criterion; the lower-bound and coherence criteria remain unmet independently of that cell.

\subsection{Artifact integrity}

Each released study package contains the executable notebook, source bundle, preregistration or frozen panel, runtime logs, raw or recovery archive, frozen CPU analysis, and a machine-readable release audit. The paper release includes a claim ledger mapping every result to its licensed wording. Large model weights and third-party datasets are not redistributed.

\section{Statement and assumption ledger}
\label{app:claim-ledger}

\subsection{Licensed and unlicensed interpretations}

\small
\begin{longtable}{@{}L{0.18\linewidth}L{0.36\linewidth}L{0.36\linewidth}@{}}
\caption{Interpretive boundaries.}\label{tab:wording}\\
\toprule
Topic & Licensed interpretation & Not licensed \\
\midrule
\endfirsthead
\toprule
Topic & Licensed interpretation & Not licensed \\
\midrule
\endhead
Predictive state & Present behavior is not a sufficient statistic for declared future-learning responses. & A finite fingerprint recovers the complete physical neural state. \\
Global fiber & The predictive quotient functor has a canonical set-level category of elements; regular bundles are conditional. & The measured Transformer system is unconditionally a smooth fiber bundle. \\
Completion & Fresh response contains visible, reuse, and irreducible-new sectors relative to a frozen projector and metrics. & The irreducible rank is the true neural hidden dimension. \\
Transport & Quotient transport is canonical; empirical finite linear transport is support- and resolution-conditioned. & The negative path study proves path independence or identity monodromy. \\
Switching & The controlled hard-ReLU system supports the proposed Hodge--$3/2$ mechanism and second-order approach while missing its full finite-grid conjunction. & Native Qwen or Mistral training universally obeys a $3/2$ law. \\
Adam & An exact algebraic matched section exists; accessible Mistral sections show future-gradient-conditioned revelation. & Generic natural reachability or a bank-invariant revelation direction. \\
\bottomrule
\end{longtable}
\normalsize

\subsection{Minimal assumption map}

\small
\begin{longtable}{@{}L{0.27\linewidth}L{0.33\linewidth}L{0.32\linewidth}@{}}
\caption{Dependencies of the principal results.}\label{tab:assumption-map}\\
\toprule
Result & Required & Not required \\
\midrule
\endfirsthead
\toprule
Result & Required & Not required \\
\midrule
\endhead
Predictive quotient and global opfibration & Complete composition and prefix-compatible probes & Topology, probability, smoothness \\
Prefix-depth filtration & Declared prefix action readouts & Differentiability \\
Visible-relative completion & Finite-dimensional Hilbert metrics and a declared visible projector & Global bundle \\
History bridge & History chart and covariance & Stationarity \\
Hodge interaction isolation & Finite task graph and common-affine memory channel & Switching assumptions \\
Graph-projected $3/2$ law & Crossing covariance, centering, and aggregate dependence conditions & Transformer or ReLU by name \\
Adam matched section & Exact Adam convention & Neural reachability \\
Neural accessibility & Finite-dimensional admissible controls and submersion & A fixed raw Euclidean conditioning threshold \\
Metric-stable radius & Declared metrics and a Lipschitz derivative bound & Coordinate-free raw units \\
Supported transport & Resolved source support and spectral gap & Ambient invertibility \\
\bottomrule
\end{longtable}
\normalsize

\subsection{Status of the mathematical package}

\begin{itemize}
\item The finite predictive core and complete-state Transformer instantiation are theorem-level results.
\item Completion, Hodge, switching, Adam, and regular transport results are conservative extensions under explicit assumptions.
\item Empirical memberships are evidence about frozen execution contexts, not axioms.
\item The resolution refinements preserve the original preregistered empirical outcomes.
\item The manuscript has undergone internal adversarial review but not external peer review or mechanized proof certification.
\end{itemize}

\phantomsection
\section*{References}
\addcontentsline{toc}{section}{References}
\renewcommand{\bibsection}{}
\setlength{\bibsep}{0.55pt plus 0.2pt minus 0.1pt}
\fontsize{8.35}{9.7}\selectfont
\raggedright
\bibliographystyle{plainnat}
\bibliography{references}
\end{document}